\documentclass[journal]{IEEEtran}

\ifCLASSINFOpdf

\fi

\usepackage{cite}
\usepackage{url}
\usepackage{amssymb}
\usepackage{pifont}
\usepackage{bbding}
\usepackage{makecell}
\usepackage{subfig}
\usepackage{graphicx}
\usepackage{multirow}
\usepackage{float}
\usepackage{diagbox}[2011/11/22]
\usepackage{epsfig}
\usepackage{threeparttable}
\usepackage{color}
\usepackage{hyperref}
\usepackage{multirow}
\usepackage{bm}
\usepackage{soul}
\usepackage{algorithm}

\usepackage{algpseudocode}
\usepackage{amsmath}

\begin{document}

\title{Kernelized Multiview Subspace Analysis by Self-weighted Learning}
\author{Huibing~Wang,
	Yang~Wang*,
    Zhao~Zhang*,
	Xianping~Fu,
	Li~Zhuo,
	Mingliang~Xu,
    Meng~Wang

\thanks{Huibing Wang and Xianping Fu are with the College of Information and Science Technology, Dalian Maritime University, Dalian, Liaoning 116021, China (e-mail: huibing.wang@dlmu.edu.cn; fxp@dlmu.edu.cn).}
\thanks{Yang Wang, Zhao Zhang, and Meng Wang are with the Key Laboratory of Knowledge Engineering with Big Data (Hefei University of Technology), Ministry of Education, Hefei University of Technology, China (e-mail: yangwang@hfut.edu.cn; cszzhang@gmail.com; eric.mengwang@gmail.com).}

\thanks{Li Zhuo is with the Faculty of Information Technology, Beijing University of Technology, Beijing, 100000, China (e-mail: zhuoli@bjut.edu.cn).} 
\thanks{Mingliang Xu is with the School of Information Engineering, Zhengzhou University, Zhengzhou 450001, China (e-mail: iexumingliang@zzu.edu.cn).}

\thanks{* Yang Wang and Zhao Zhang are corresponding authors.}
}

\markboth{IEEE Transactions on Multimedia}%
{Shell \MakeLowercase{\textit{et al.}}: Bare Demo of IEEEtran.cls for IEEE Journals}

\maketitle

\begin{abstract}
With the popularity of multimedia technology, information is always represented or transmitted from multiple views. Features from multiple views are combined into multiview data. Even though multiview data can reflect the same sample from different perspectives, multiple views are consistent to some extent because they are representations of the same sample. Most of the existing algorithms are graph-based ones to learn the complex structures within multiview data but overlook the information within data representations. Furthermore, many existing works treat multiple views discriminatively by introducing some hyperparameters, which is undesirable in practice. To this end, abundant multiview-based methods have been proposed for dimension reduction. However, there is still no research that leverages the existing work into a unified framework. To address this issue, in this paper, we propose a general framework for multiview data dimension reduction, named kernelized multiview subspace analysis (KMSA). It directly handles the multiview feature representation in the kernel space, providing a feasible channel for the direct manipulation of multiview data with different dimensions. In addition, compared with the graph-based methods, KMSA can fully exploit information from multiview data with nothing to lose. Furthermore, since different views have different influences on KMSA, we propose a self-weighted strategy to treat different views discriminatively according to their contributions. A co-regularized term is proposed to promote the mutual learning from multiviews. KMSA combines self-weighted learning with the co-regularized term to learn the appropriate weights for all views. We also discuss the influence of the parameters in KMSA regarding the weights of the multiviews. We evaluate our proposed framework on 6 multiview datasets for classification and image retrieval. The experimental results validate the advantages of our proposed method.
\end{abstract}


\IEEEpeerreviewmaketitle

\section{Introduction}

\begin{table}[htbp]
	\center
	\caption{Summarizations of typical multiview DR algorithms: ``Data driven'' means that multiview data (and not just the graph) participate in the subspace construction process. ``Self-weighted learning'' means that the algorithm can automatically learn the weights for all views. ``Framework'' means that the algorithm can be utilized as a generalized framework to extend some other single-view methods into the multiview mode. The comparison methods include multiview dimensionality co-reduction (MDcR) \cite{zhang2017flexible}, multiview spectral embedding (MSE) \cite{Xia2010}, generalized multiview analysis (GMA) \cite{sharma2012generalized}, canonical correlation analysis (CCA) \cite{michaeli2016nonparametric}, multiview discriminant analysis (MvDA) \cite{kan2016multi}, and the co-regularized approach (Co-Regu) \cite{kumar2011co}.}
	\begin{tabular}
		{lcccc}
		\hline
		& Data driven & Self-weighted learning & Framework \\
		\hline
		
		MDcR \cite{zhang2017flexible}  & \Checkmark &\Checkmark &\ding{56} \\
		MSE \cite{Xia2010}  & \ding{56}&\Checkmark &\ding{56} \\
		GMA \cite{sharma2012generalized} &\Checkmark & \ding{56}&\Checkmark \\
		CCA \cite{michaeli2016nonparametric} &\Checkmark &\ding{56} &\ding{56} \\
		MvDA \cite{kan2016multi} & \Checkmark& \ding{56}& \ding{56}\\
		Co-Regu \cite{kumar2011co}  &\ding{56} & \ding{56}& \ding{56}\\
		\textbf{KMSA (Ours)}  &\Checkmark &\Checkmark & \Checkmark\\
		\hline
	\end{tabular}
	\label{tab1}
\end{table}

\IEEEPARstart{W}{ith} the development of information technology, we have witnessed a surge of techniques to describe the same sample from multiple views \cite{Xia2010,guo2018partial,LiG2012,wang2017unsupervised,wang2020surveyTOMM}. Multiview data generated from various descriptors \cite{wei2018glad} or sensors are commonly seen in real-world applications \cite{XuW2018,cao2015diversity,zhang2017latent}, which has hastened the related research on multiview learning \cite{zhang2016deep}. For example, one image can always be represented by different descriptors, such as local binary patterns (LBPs) \cite{Ojala2002}, the scale-invariant feature transform (SIFT) \cite{rublee2011orb}, histograms \cite{dalal2005histograms} and locality-constrained linear coding (LLC) \cite{wang2010locality}. For text analysis \cite{dong2018predicting}, documents can be written in different languages \cite{bisson2012co}. Notably, multiview data may share consistent correlation information \cite{wang2015robust,wang2016iterative,wang2018multiview}, which is crucial to promote the performance of related tasks \cite{wang2020survey,li2002statistical,zhang2018generalized,liu2018late}.

\begin{figure*}[ht]
	\centerline{\includegraphics[width=6.8in]{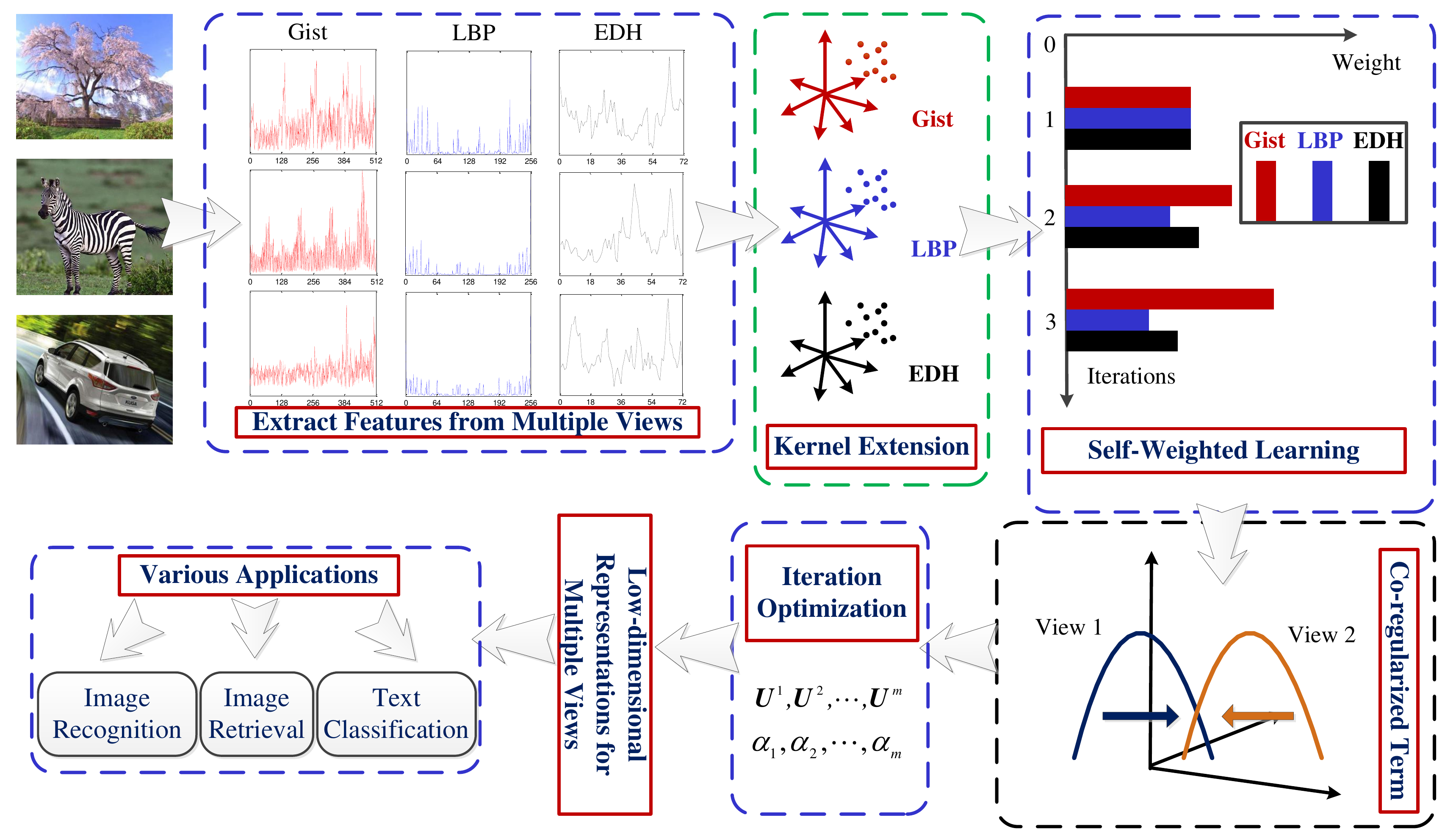}}
	\caption{The flow chart of kernelized multiview subspace analysis (KMSA), which handles multiview data representations within the kernel space. KMSA adaptively learns the weights for multiple views.
A co-regularized term is proposed to minimize the divergence of different views. Finally, an iterative optimization process is proposed to jointly learn the low-dimensional subspace of multiview data and the view-wise weight parameters. (This figure is best viewed in color.)  }
	\label{fig1}
\end{figure*}

Multiview dimensional reduction (DR) methods have been well studied in many applications \cite{wu2018and,nie2018auto,nie2018multiview}. In particular, Kumar et al. \cite{kumar2011co} proposed a multiview spectral embedding approach by introducing a co-regularized framework that can narrow down the divergence between graphs from multiple views. Xia et al. \cite{Xia2010} introduced an autoweighted method to construct common low-dimensional representations for multiple views, which has achieved good performances in image retrieval and clustering. Wang et al. \cite{wang2018multiview} exploited the consensus of multiview structures beyond the low rankness to construct low-dimensional representations for multiview data to boost the clustering performance. Kan et al. \cite{kan2016multi} extended linear discriminant analysis (LDA) \cite{izenman2013linear} to multiview discriminant analysis (MvDA), which updates the projection matrices for all views through an iterative procedure. Luo et al. \cite{luo2015tensor} proposed a tensor canonical correlation analysis (TCCA) to address multiview data in the general tensor form. TCCA is an extension of CCA \cite{michaeli2016nonparametric} and has achieved ideal performances in many applications. Zhang et al. \cite{zhang2017flexible} proposed a novel method to flexibly exploit the complementary information between multiple views on the stage of dimension reduction while preserving the similarity of data points across different views. Self-weighted multiple kernel learning (SMKL) \cite{kang2018self} utilizes a self-weighting scheme to learn a new kernel matrix by combining multiple kernels. Therefore, it is different from the multiview learning methods and cannot construct a low-dimensional subspace for the original multiview data. Furthermore, some generative adversarial network (GAN)-based \cite{goodfellow2014generative} methods \cite{zhu2018visual} can also generate a low-dimensional representation for multiview data.

Presently, most of the multiview DR methods \cite{kumar2011co,Xia2010,nie2018auto
} are graph-based approaches \cite{cui2017general} that care more about data correlations and overlook information regarding multiview data. Likewise, these limitations hold for numerous studies \cite{kumar2011co,nie2018auto}.
The following are a few typical works: Multiview spectral embedding (MSE) \cite{Xia2010} is an extension of Laplacian eigenmaps (LE) \cite{belkin2002laplacian} and considers the Laplacian graphs between multiview data rather than the information within the data representation. Kumar et al. \cite{kumar2011co} also exploited only the information within the Laplacian graphs and utilized a co-regularized term to minimize the divergence between different views. However, this method failed to exploit the information within a multiview data representation. Even though there are some approaches, such as MvDA \cite{izenman2013linear}, CCA \cite{michaeli2016nonparametric}, etc., that can fully consider the original multiview data and extend traditional DR \cite{mika1999fisher} to the multiview version, these failed to provide a general framework for most DR approaches. Therefore, how to construct a general framework to integrate features from multiple views to construct low-dimensional representations while achieving the ideal performance is the goal.

In this paper, we aim to develop a unified framework to project multiview data into a low-dimensional subspace. Our proposed kernelized multiview subspace analysis (KMSA) is equipped with a self-weighted learning method to make different weights for multiple views according to their contributions. We also discuss the influence of the parameter $r$ in KMSA for the learned weights of multiple views in $\bm{\alpha}$. Furthermore, KMSA adopts the co-regularized term to minimize the divergence between every two views, which can encourage all views to learn from each other. The construction process of KMSA is shown in Fig. \ref{fig1}. We compare KMSA with some typical methods in Table \ref{tab1}.

We remark that Yan et al. \cite{yan2007graph} proposed a framework for dimensional reduction techniques. Different from that, KMSA extends the framework to kernel space with multiviews to address the problems that are caused by different dimensions of features from multiple views. Then, KMSA adopts a self-weighted learning technique to add different weights to these views according to their contributions. Finally, KMSA is equipped with a co-regularized term to minimize the divergence between different views to achieve multiview consensus.
\begin{table}[htp]
	\renewcommand{\arraystretch}{1.3}
	\caption{The descriptions of some important formula symbols}
	\begin{tabular}
		{l|l}
		\Xhline{1.2pt}
		\textbf{Notation}&  \textbf{Description}\\
		\hline
		\hline
		$\bm{X}^v \in \mathbb{R}^{{D_v}\times N}$& set of all features in the $v$th view\\
		
		$\bm{Y}^v \in \mathbb{R}^{{d}\times N}$& set of low-dimensional representations in the $v$th view\\
		
		$\bm{x}_i^v \in \mathbb{R}^{{D_v}\times 1}$ & the $i$th feature in the $v$th view\\
		
		$\bm{y}_i^v\in \mathbb{R}^{{d}\times N}$ & low-dimensional representation for $\bm{x}_i^v$\\
		
		$D_v\in \mathbb{R}^{1}$ & dimension of the features in the $v$th view\\
		
		$\bm{s}_i^v\in \mathbb{R}^{{N}\times 1}$ & sparse relationships for the $i$th feature in the $v$th view\\
		
		$\bm{w}^v \in \mathbb{R}^{{D_v}\times 1}$ & projection direction for the $v$th view\\
		
		$\bm{S}^v\in \mathbb{R}^{{N}\times N}$ & sparse reconstructive matrix for features in the $v$th view\\
		
		$\bm{K}^v\in \mathbb{R}^{{N}\times N}$ & kernel matrix for features in the $v$th view\\
		
		$\bm{U}^v\in \mathbb{R}^{{N}\times d}$ & coefficient matrix for the $v$th view\\
		
		$\bm{u}^v\in \mathbb{R}^{{N}\times 1}$ & coefficient vector for the $v$th view\\
		
		$\alpha_v\in \mathbb{R}^{1}$ & weighting factor for the $v$th view\\
		
		$r\in \mathbb{R}^{1}$ & power exponent for the weights $\alpha_1, \alpha_2, \cdots, \alpha_v$\\
		
		$\bm{\mathcal{Q}}^v\in \mathbb{R}^{{D_v}\times D_v}$ & constraint matrix for the $v$th view \\
		
		\Xhline{1.2pt}
	\end{tabular}
	\label{tab2}
\end{table}
The major contributions of this paper are summarized as follows:

\begin{itemize}
    \item We developed a novel framework named KMSA for the task of multiview dimension reduction. We discussed that most of the eigen-decomposition-based DR methods \cite{jolliffe2011principal,he2004locality} can be extended to the corresponding multiview versions throughout KMSA.
    \item KMSA fully considers both the single-view graph correlations between multiple views to calculate the importance of all views, which is an attempt to combine self-weighted learning with a co-regularized term to deeply exploit the information from multiview data.
    \item  We discussed the details of the optimization process for KMSA, with the results showing that KMSA can achieve a state-of-the-art performance.
\end{itemize}

\section{Kernel-based Multiview Embedding with Self-weighted Learning}

In this section, we discuss the intuition of our proposed KMSA method.

Assume that we are given a multiview dataset $\bm{X} = \left\{ \bm{X}^v \in \mathbb{R}^{D_v \times N}, v=1,\cdots,m \right\}$, which consists of $N$ samples from $m$ views, where $\bm{X}^v \in \mathbb{R}^{D_v \times N}$ contains all features from the $v$th view. $D_v$ is the dimensions of features from the $v$th view. $N$ is the number of training samples. The goal of KMSA is to construct an appropriate architecture to obtain low-dimensional representations $\bm{Y} = \left\{ \bm{Y}^v \in \mathbb{R}^{d \times N} \right\}$ for the original multiview data, where $d<D_v, v=1,\cdots,m$. The notations utilized in this paper are summarized in Table \ref{tab2}.

\subsubsection{\textbf{Kernelization for Single-view Data}}
The proposed KMSA extension of the single-view DR method is divided into kernel spaces, which provides a feasible way to conduct direct manipulations on multiview data rather than similarity graphs. Before taking the kernel space into consideration, KMSA exploits the heterogeneous information for each view as follows:

\begin{equation}
\label{eq1}
\begin{array}{l}
\phi^v = \mathop {\min }\limits_{\bm{w}^v} \sum\limits_{i \neq j}
{\left\| {\left( {\bm{w}^v} \right)^T\bm{x}_i^v - \left( \bm{w}^v\right)^T\bm{x}_j^v} \right\|}^2 \bm{S}_{ij}^v \\
 s.t.\;\left( {\bm{w}^v} \right)^T\bm{\mathcal{Q}}^v
 \bm{w}^v = 1,\\
 \end{array}
\end{equation}
where $\bm{w}^v \in \mathbb{R}^{D_v \times 1}$ is the projection vector. $\bm{S}_{ij}^v$ is the correlation between $\bm{x}_i^v$ and $\bm{x}_j^v$ in the $v$th view. $\bm{\mathcal{Q}}^v = \bm{X}^v\bm{B}^v\left({\bm{X}^v} \right)^T$ or $\bm{\mathcal{Q}}^v = \bm{I}$ according to their respective different constraints of various dimensional reduction algorithms. Most algorithms can be generated automatically by using different construction tricks of $\bm{S}^v$ and $\bm{\mathcal{Q}}^v$, which has been illustrated in \cite{yan2007graph}. $\phi^v$ can be further expressed as $ \phi^v = \left( \bm{w}^v \right)^T\bm{X}^v \bm{P}^v \left({\bm{X}^v} \right)^T\bm{w}^v$ according to the mathematical transformation \cite{yan2007graph} and $ \bm{P}^v =\bm{E}^v - \bm{S}^v$, where $\bm{E}^v$ is the diagonal matrix and $\bm{E}_{ii}^v = \sum_{j\neq i}\bm{S}_{ij}^v$. To facilitate KMSA in addressing multiview data, we project all feature representations into kernel space as $\varphi :\mathbb{R}^{D_v } \to \digamma, \bm{x}_i^v \to \varphi \left({\bm{x}_i^v}  \right)$. $\varphi$ is a nonlinear mapping function. $\bm{X}_\varphi ^v = \left[ {\varphi \left( {\bm{x}_1^v} \right),\varphi \left({\bm{x}_2^v } \right), \cdots ,\varphi \left( {\bm{x}_N^v } \right)} \right]$ contains features that have been mapped into the kernel space $\digamma$.

Then, we extend Eq. \ref{eq1} into the kernel representation as follows:
\begin{equation}
\label{eq2}
\begin{array}{l}
 \mathop {\min }\limits_{\bm{w}_\varphi ^v} \left( {\bm{w}_\varphi^v}  \right)^T\bm{X}_\varphi ^v \bm{P}^v \left( {\bm{X}_\varphi ^v } \right)^T\bm{w}_\varphi ^v \\
 s.t.\;\;\;\left( {\bm{w}_\varphi ^v} \right)^T\bm{\mathcal{Q}}^v\bm{w}_\varphi ^v = 1 ,\\
 \end{array}
\end{equation}
where $\bm{w}_\varphi^v$ is the projection direction of $\bm{X}_\varphi ^v$ and is located in the space spanned by ${\varphi \left( {\bm{x}_1^v} \right),\varphi \left({\bm{x}_2^v } \right), \cdots ,\varphi \left( {\bm{x}_N^v } \right)}$. Consequently, $\bm{w}_\varphi^v$ can be replaced with $\bm{w}_\varphi^v = \sum\limits_{i = 1}^m {\alpha _i^v \varphi \left( {\bm{x}_i^v } \right)} = \bm{X}_\varphi ^v \bm{u}^v$. Then, Eq. \ref{eq2} can be further modified as follows:

\begin{equation}
\label{eq3}
\begin{array}{l}
 \mathop {\min }\limits_{\bm{u}^v} \left( {\bm{u}^v}\right)^T\bm{K}^v\bm{P}^v\bm{K}^v\bm{u}^v \\
 s.t.\;\;\;\left( {\bm{u}^v} \right)^T\bm{\mathcal{M}}^v\bm{u}^v = 1. \\
 \end{array}
\end{equation}
$\bm{K}^v = \left( {\bm{X}_\varphi^v} \right)^T\bm{X}_\varphi^v \in \mathbb{R}^{N\times N}$ is the kernel matrix, which is symmetric, and $\bm{K}_{ij}^v = \left( {\varphi \left( {\bm{x}_i^v} \right) \cdot \varphi \left( {\bm{x}_j^v } \right)} \right)$. $\bm{\mathcal{M}}^v = \bm{K}^v\bm{B}^v\bm{K}^v$ or $\bm{\mathcal{M}}^v = \bm{K}^v$, which corresponds to the setting of $\bm{\mathcal{Q}}^v$. Therefore, if we want to obtain an optimal subspace with $d$ dimensions, $\bm{u}_1^v ,\bm{u}_2^v, \cdots ,\bm{u}_d^v$ can be utilized to construct the subspace corresponding to the largest $d$ positive eigenvalues of $\bm{K}^v\bm{P}^v\bm{K}^v$, which is equivalent to finding the coefficient matrix $\bm{U}^v = \left[ {\bm{u}_1^v ,\bm{u}_2^v, \cdots ,\bm{u}_d^v} \right] \in \mathbb{R}^{N\times d}$ as follows:

\begin{equation}
\label{eq4}
\begin{array}{l}
 \mathop {\min }\limits_{\bm{U}^{\left( v \right)}} tr\left\{ {\left( {\bm{U}^v} \right)^T\bm{K}^v\bm{P}^v\bm{K}^v\bm{U}^v} \right\} \\
 s.t.\;\;\left( {\bm{U}^v} \right)^T\bm{\mathcal{M}}^v\;\bm{U}^v = \bm{I} .\\
 \end{array}
\end{equation}
The low-dimensional representations of the original $\bm{X}^v$ are $\bm{Y}^v = \left[ {\bm{y}_1^v ,\bm{y}_2^v, \cdots ,\bm{y}_N^v} \right] = \left( {\bm{X}_\varphi ^v} \bm{U}^v\right)^T\bm{X}_\varphi ^v = \left( {\bm{U}^v} \right)^T\bm{K}^v \in \mathbb{R}^{d\times N}$. Even though we can extend DR methods into the kernel space to avoid the problem where the dimensions of features from multiple views are different from each other, the construction procedures of $\left\{\bm{Y}^v, v=1,\cdots,m\right\}$ are still independent and waste much information from the other views.
\begin{figure}[htbp]
	\centerline{\includegraphics[width=3.3in]{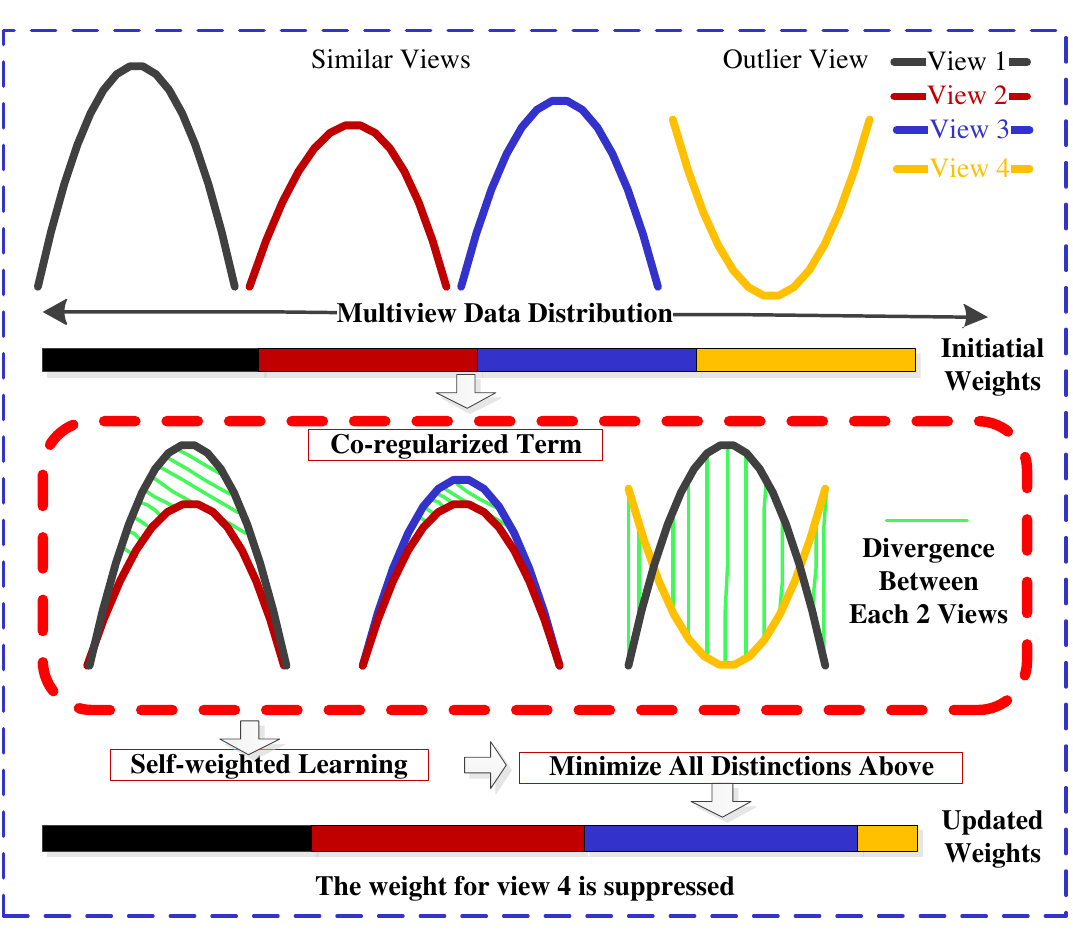}}
	\caption{The learning process of $\alpha_1,\alpha_2,\cdots,\alpha_m$ via self-weighted learning and a co-regularized term. Because the distribution of data in the 4th view is different from the distributions of the 3 other views, the divergence between the 4th view and the other views will be large. The self-weighted learning procedure gives the 4th view a smaller weight to minimize the co-regularized term. (This figure is best viewed in color.)}
	\label{fig2}
\end{figure}

\subsubsection{\textbf{Self-weighted Learning of the Weights for Multiple Views}}
To integrate information from multiple views, the most straightforward way is to minimize the sum of Eq. \ref{eq4} for all $m$ views. Then, we can obtain the following objective function:

\begin{equation}
\label{eq5}
\begin{array}{l}
 \mathop {\min }\limits_{\bm{U}^1,\bm{U}^2, \cdots,\bm{U}^m} \sum\limits_{v = 1}^m {tr\left\{ {\left( {\bm{U}^v} \right)^T\bm{K}^v\bm{P}^v\bm{K}^v\bm{U}^v} \right\}} \\
 s.t.\;\;\;\left( {\bm{U}^v} \right)^T\bm{\mathcal{M}}^v\bm{U}^{\left( v \right)} = I,\;\forall 1 \le v \le m .\\
 \end{array}
\end{equation}
However, different views make different contributions to the objective value in Eq. \ref{eq5}. Some adversarial views may make a negative contribution to the final low-dimensional representations. Therefore, it is rational to treat these views discriminatively. We propose different weighting factors for these views while refining the low-dimensional representations. Therefore, the self-weighted learning strategy is as follows:
\begin{equation}
\label{eq6}
\begin{array}{l}
 \mathop {\min }\limits_{\bm{\alpha},\bm{U}^1,\bm{U}^2, \cdots,\bm{U}^m} \sum\limits_{v = 1}^m {\alpha_v^r tr\left\{ {\left( {\bm{U}^v} \right)^T\bm{K}^v\bm{P}^v\bm{K}^v\bm{U}^v} \right\}} + \kappa \|\bm{\alpha}\|_r^r \\
 s.t.\;\;\;\left( {\bm{U}^v} \right)^T\bm{\mathcal{M}}^v\bm{U}^{\left( v \right)} = I,\;\forall 1 \le v \le m \\
 \;\;\;\;\;\;\;\; \sum\limits_{v = 1}^m{\alpha_v} = 1, r>1,
 \end{array}
\end{equation}
where $\bm{\alpha} = \left[ \alpha_1, \alpha_2,\cdots,\alpha_m \right]$. $\kappa$ is a trade-off between the two terms mentioned above. The second term in Eq. 6 aims to make all values in $\bm{\alpha}$ nonnegative, which can force all views to participate in the process of multiview learning. A larger $\kappa$ will lead Eq. 6 to be more inclined toward the second term. $r>1$ ensures that all views make particular contributions to the final low-dimensional representations $\left\{\bm{Y}^v,v=1,2,\cdots,m\right\}$. Otherwise, only one entry in $\bm{\alpha}$ will be $1$, while the other $m-1$ entries will be zero. The second term in Eq. 6 minimizes the $r$th power of the $\ell$-$r$ norm for $\bm{\alpha}$, which can also make $\bm{\alpha}$ as nonsparse as possible. The rationale is that $\|\bm{\alpha}\|_r^r = \sum_{v = 1}^m {\alpha_v^r}$ achieves its minimum when $\alpha_v = 1/m$ with respect to $\sum_{v = 1}^m {\alpha} =1$. Therefore, the second term in Eq. 6 can further promote the participation for all views. A larger $r$ will cause all weights $\alpha_v \left(v=1,2,\cdots,m\right)$ to be similar to each other. These two techniques can equip these views with different weights according to their contributions.

The intuition of our self-weighted scheme is as follows: for the $v$th view in Eq. \ref{eq6}, its optimal solution $\bm{U}^v$ can be obtained by minimizing the trace of its corresponding term. However, considering all views, the values of some traces may be large due to the unsatisfactory relationships between features from corresponding views, which also causes the obtained $\bm{U}$ to be unsatisfactory. Therefore, it is obvious that if the trace of one view is large, the information maintained in the view is less important. Smaller weights should be assigned to these views because the sum of all views equals to 1.

According to Eq. \ref{eq6}, we can obtain the low-dimensional representations $\left\{\bm{Y}^v,v=1,2,\cdots,m\right\}$ simultaneously. However, the construction process of each $\bm{Y}^v$ cannot learn from the information from the other views. Even though we have set different views with different weights, the learned $\left\{\bm{Y}^v,v=1,2,\cdots,m\right\}$ are equal to those in Eq. \ref{eq4}. Finally, we propose a co-regularized term to help all views to learn from each other.

\subsubsection{\textbf{Minimize the Divergence between Different Views by a Co-regularized Term}}
Multiview learning aims to enable all views to learn from each other to improve the overall performance; hence, it is essential for KMSA to develop a method to integrate compatible and complementary information from all views. Some researchers \cite{kumar2011co} have attempted to minimize the divergence between low-dimensional representations via various co-regularized terms, which can facilitate the transfer of information across views.

Because the coefficient matrix $\bm{U}^v$ is used to reconstruct the low-dimensional representations, each column of $\bm{U}^v$ can be regarded as a coding of the original samples. Therefore, KMSA attempts to minimize the divergence between the two coefficient matrices from each pair of views as follows:

\begin{equation}
\label{eq7}
D\left( {\bm{U}^i,\bm{U}^j } \right) = \left\| \frac{{\bm{L}_{\bm{U}^i}}}{\left\|{\bm{L}_{\bm{U}^i}}\right\|_F^2} - \frac{{\bm{L}_{\bm{U}^j}}}{\left\|{\bm{L}_{\bm{U}^j}}\right\|_F^2} \right\|_F^2.
\end{equation}
We define $\bm{L}_{\bm{U}^i} = \left(\bm{U}^i\right)^T\bm{U}^i$, and $\bm{L}_{\bm{U}^i}$ is a graph that contains the relationships between all features in the $i$th view. The $r$th row with the $c$th column element in $\bm{L}_{\bm{U}^i}$ is equal to $\left(\bm{u}_r^i\right)^T\bm{u}_c^i$. Therefore, $\bm{L}_{\bm{U}^i}$ is an adjacency matrix that is typically a linear kernel matrix. It is a graph whose nodes are features from the $i$th view and whose edge weights are calculated by taking the inner product of every 2 features. Minimizing Eq. \ref{eq7} encourages every two views to learn from each other and bridge the gap between them. Furthermore, $D\left( {\bm{U}^i,\bm{U}^j } \right)$ can be replaced with $ - tr\left\{ {\left( {\bm{U}^i} \right)^T\bm{U}^i\left( {\bm{U}^j } \right)^T\bm{U}^j } \right\}$ through mathematical deductions \cite{kumar2011co}. We utilize Eq. \ref{eq7} as one regularized term in KMSA in the following content.

\subsubsection{\textbf{Overall Objective Function}}

Based on the above, we propose the following objective function:
\begin{equation}
\label{eq8}
\begin{array}{l}
 \min \mathcal{G}\left(\bm{\alpha},\bm{U}^1,\bm{U}^2, \cdots,\bm{U}^m\right) = \\
~~~~~\sum\limits_{v = 1}^m {\alpha_v^r \underbrace{ tr\left\{ {\left( {\bm{U}^v} \right)^T\bm{K}^v\bm{P}^v\bm{K}^v\bm{U}^v} \right\}}_{\text{Kernel-based subspace learning}}} + \underbrace{\kappa \|\bm{\alpha}\|_r^r}_{\text{regularization }} \\
~~~~~+\underbrace{\sum\limits_{v\neq w} \frac{\alpha_v^r+\alpha_w^r}{2\eta}tr\left\{ {\left( {\bm{U}^v} \right)^T\bm{U}^v\left( {\bm{U}^w } \right)^T\bm{U}^w } \right\}}_{\text{Minimize divergence between $\bm{U}^v $ and $\bm{U}^w$s }}\\
 s.t.\;\;\;\left( {\bm{U}^v} \right)^T\bm{\mathcal{M}}^v\bm{U}^{\left( v \right)} = I,\;\forall 1 \le v \le m \\
 \;\;\;\;\;\;\;\; \sum\limits_{v = 1}^m{\alpha_v} = 1, r>1,
 \end{array}
\end{equation}
where $\eta$ is a negative constant. The co-regularized term between the $v$th and $w$th views in Eq. \ref{eq8} should take into consideration the importance of these 2 views. Because $\alpha_v^r$ and $\alpha_w^r$ can reflect the importance of the $v$th and $w$th view, the weight factor of the co-regularized term should combine these $2$ weight factors into one form. Therefore, we assign the weight factor of the co-regularization term by modifying the mean value of $\alpha_v^r$ and $\alpha_w^r$ as $\frac{\alpha_v^r+\alpha_w^r}{2\eta}$. $\eta$ is a negative constant used to adjust the numerical scale of $\frac{\alpha_v^r+\alpha_w^r}{2}$. Furthermore, $\eta$ is able to control the strength to minimize the divergence between different views. The larger the value of $|\eta|$ is, the smaller the influence of the co-regularized term. It is notable that $\alpha_v^r$ and $\alpha_w^r$ are learned automatically by considering both the graph for each view and the correlations of multiple views, and $\alpha_i^r, i=1,2,\cdots,m$ can obtain better solutions. It has the following 2 advantages:

\begin{itemize}
	\item  $\left(\alpha_v^r+\alpha_w^r\right)/2\eta$ can better reflect the influence of the regularized term between these two views. Compared with KMSA, some multiview learning methods \cite{kumar2011co} have $m$ parameters to set. This matter could become even worse as the number of views increases. Fortunately, only one parameter $\eta$ needs to be set for KMSA, which can better balance the influence of the co-regularized term.
	
	\item The learning process of $\alpha_1,\alpha_2,\cdots,\alpha_m$ fully considers the correlations between different views. Minimizing Eq. \ref{eq8} means that some similar views receive larger weights, and the obtained low-dimensional representations are inclined to be consistent views while avoiding the disturbance of some adversarial views, as in Fig. \ref{fig2}.
\end{itemize}

We can obtain the low-dimensional representations for these views as $\bm{Y}^v = \left[ {\bm{y}_1^v ,\bm{y}_2^v, \cdots ,\bm{y}_N^v} \right] = \left( {\bm{U}^v\bm{X}_\varphi ^v} \right)^T\bm{X}_\varphi ^v = \left( {\bm{U}^v} \right)^T\bm{K}^v \in \mathbb{R}^{d\times N}$. $\bm{U}^v$ can be calculated by Eq. \ref{eq8} with eigenvalue decomposition.

\subsection{Optimization Process for KMSA }

In this section, we provide the optimization process for KMSA.  We develop an alternating optimization strategy, which separates the problem into several subproblems such that each subproblem is tractable. That is, we alternatively update each variable when fixing the others. We summarize the optimization process in Algorithm \ref{algorithm1}.

\begin{algorithm}[htb]
	\caption{Optimization Process for KMSA}
	\label{algorithm1}
	\begin{algorithmic}[1]
		\Require \\
		Initialize $\bm{U}^v,v=1,2,\cdots,m$ using Eq. \ref{eq4}; \\
		Initialize $\alpha_v = 1/m, v=1,2,\cdots,m$;\\
		Set the parameters $\kappa, \eta$ and $r$;
		\Ensure
		
		\State Calculate $\bm{K}^v$ and $\bm{P}^v$ using the original multiview data
		\State for $t=1:iter$
		\State 	~~~~for $i=1:m$
		\State ~~~~~~~~fix $\bm{U}^v_t, v=1,2,\cdots,i-1,i+1,\cdots,m$, update
		\State  ~~~~~~~~$\bm{U}^i_t$ using Eq. \ref{eq9}.
		\State ~~~~end $i$
		
		\State 	~~~~for $j=1:m$
		\State ~~~~~~~~fix $\alpha_v, v=1,2,\cdots,i-1,i+1,\cdots,m$ and
		\State ~~~~~~~~$\bm{U}^l_t, l=1,2,\cdots,m $, update $\alpha_j$ using Eq. \ref{eq15}.
		\State ~~~~end $j$
		\State end $t$	
		
		\State Calculate the low-dimensional representations according to Eq. \ref{eq16}.\\
		
		\Return $\bm{Y}^v,v=1,2,\cdots,m$;
	\end{algorithmic}
\end{algorithm}

\textbf{Updating $\bm{U}^v$}: By fixing all variables but $\bm{U}^v$, Eq. \ref{eq8} will reduce to the following equation without considering the constant additive and scaling terms:

\begin{equation}
\label{eq9}
\begin{array}{l}
\min \mathcal{F}\left(\bm{U}^v\right) = tr\left\{ {\left( {\bm{U}^v} \right)^T\bm{K}^v\bm{P}^v\bm{K}^v\bm{U}^v} \right\} \\
 ~~~~~~~~~~+\sum\limits_{v\neq w} \frac{1+\left(\alpha_w/\alpha_v\right)^r}{2\eta}tr\left\{ {\left( {\bm{U}^v} \right)^T\bm{U}^v\left( {\bm{U}^w } \right)^T\bm{U}^w } \right\}\\
 s.t.\;\;\;\left( {\bm{U}^v} \right)^T\bm{\mathcal{M}}^v\bm{U}^{\left( v \right)} = I,\\
 \end{array}
\end{equation}

which has a feasible solution; $\mathcal{F}\left(\bm{U}^v\right)$ can be transformed according to the operational rules of the matrix trace as follows:

\begin{equation}
\label{eq10}
\begin{array}{l}
\min \mathcal{F}\left(\bm{U}^v\right)= \\
tr\left\{ {\left( {\bm{U}^v} \right)^T\left(\bm{K}^v\bm{P}^v\bm{K}^v +\sum\limits_{v\neq w} \frac{1+\left(\alpha_w/\alpha_v\right)^r}{2\eta} \bm{U}^w\left( {\bm{U}^w } \right)^T \right){\bm{U}^v}} \right\}.\\
\end{array}
\end{equation}
We set $ \mathcal{H}^v= \bm{K}^v\bm{P}^v\bm{K}^v +\sum\limits_{v\neq w} \frac{1+\left(\alpha_w/\alpha_v\right)^r}{2\eta}\bm{U}^w\left( {\bm{U}^w } \right)^T$. Therefore, with the constraint $\left( {\bm{U}^v} \right)^T\bm{\mathcal{M}}^v\bm{U}^{\left( v \right)} = I$, the optimal $\bm{U}^v$ can be solved by generalized eigen-decomposition as $\mathcal{H}^v\bm{u} = \xi\bm{\mathcal{M}}^v\bm{u}$. $\bm{U}^v$  consists of eigenvectors that correspond to the smallest $d$ eigenvalues. $\bm{U}^1,\bm{U}^2, \cdots,\bm{U}^m$ can be calculated by the above procedure to update them.

\textbf{Updating $\bm{\alpha}$}: After $\bm{U}^1,\bm{U}^2, \cdots,\bm{U}^m$ are fixed as above, $\bm{\alpha}$ is updated. By using a Lagrange multiplier $\lambda$ to take the constraint $\sum_{v=1}^m \alpha_v = 1$ into consideration, we obtain the following Lagrange function:
\begin{equation}
\label{eq11}
\mathcal{L}\left(\bm{\alpha},\lambda\right) = \mathcal{G}\left(\bm{\alpha},\bm{U}^1,\bm{U}^2, \cdots,\bm{U}^m\right) - \lambda \left( \sum\limits_{v=1}^m \alpha_v-1 \right).
\end{equation}

\begin{figure*}[htp]
	\centerline{
		\subfloat[Corel1K]{\includegraphics[width=2.25in]{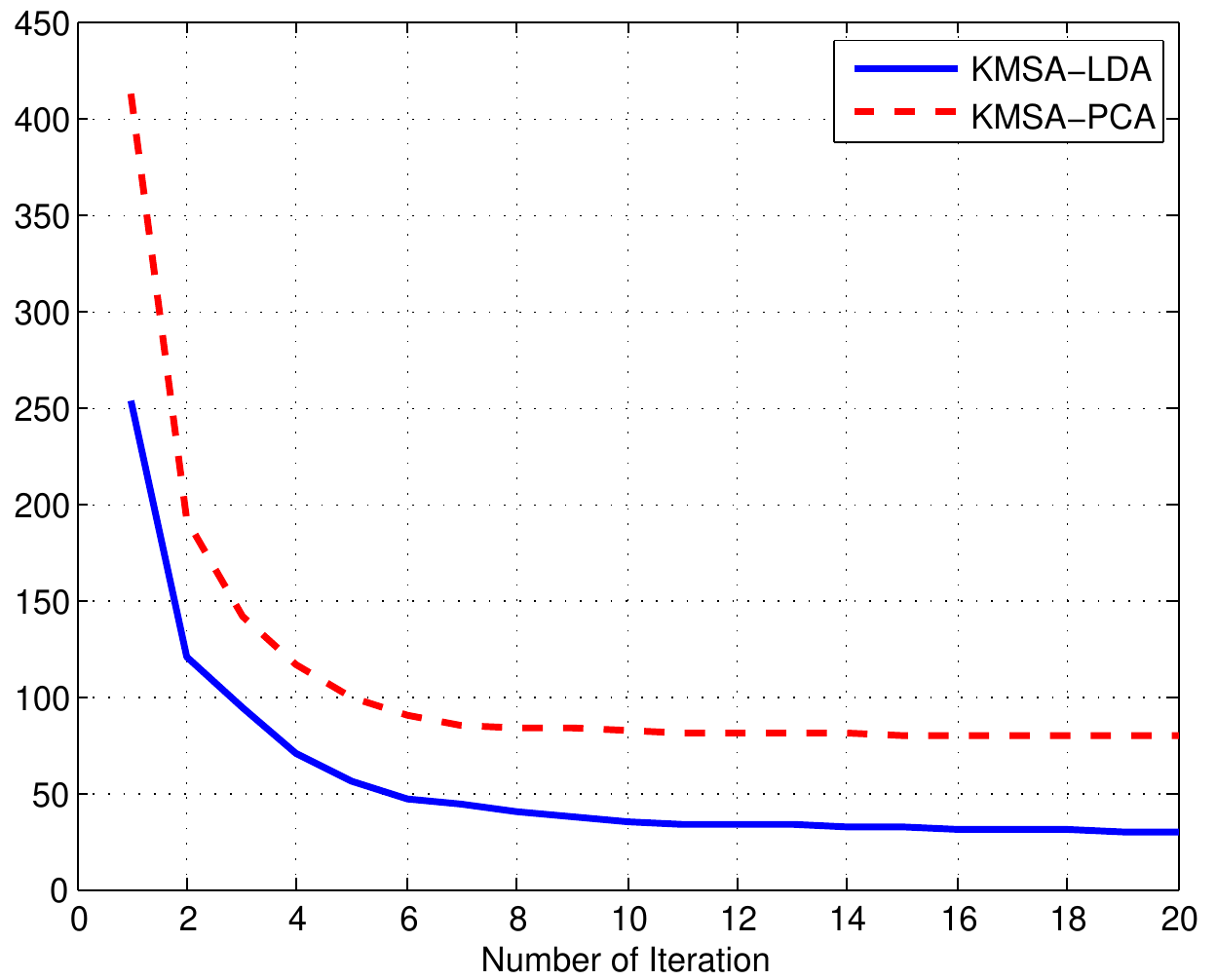}}	
		\subfloat[Caltech101]{\includegraphics[width=2.22in]{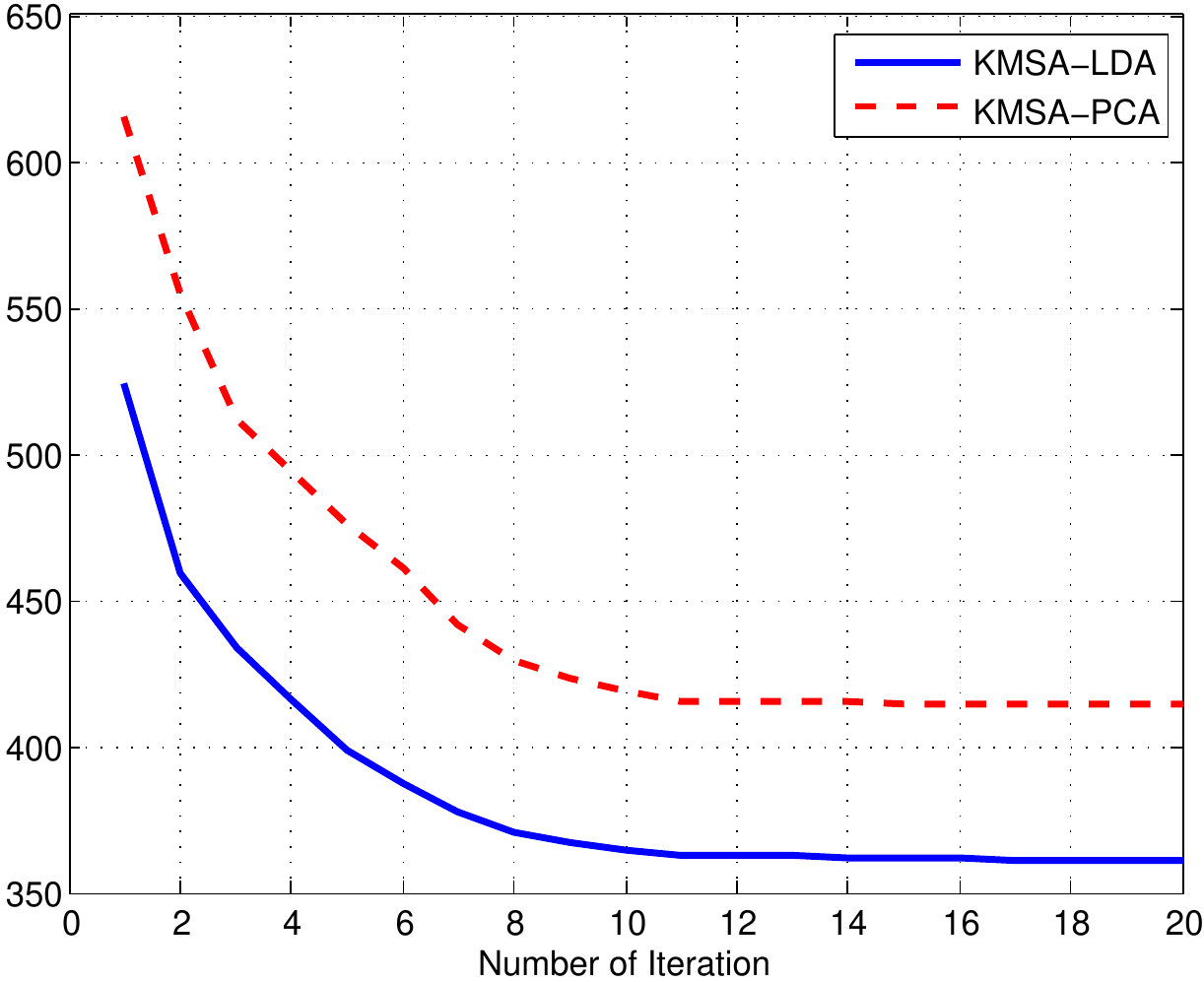}}
		\subfloat[ORL]{\includegraphics[width=2.23in]{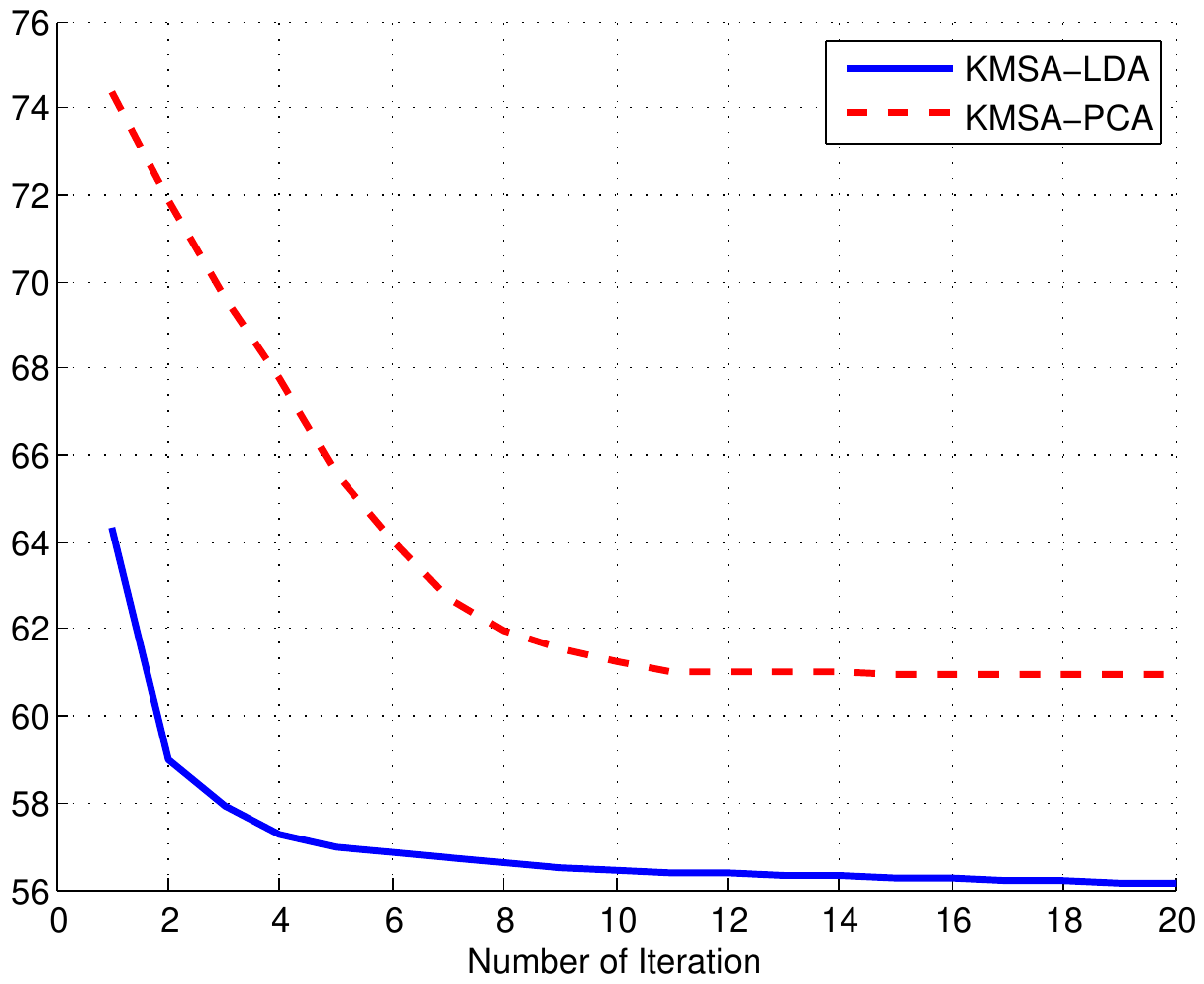}}}
	\caption{The variation of the value of the objective function with the number of iterations. The values decrease as the number of iterations increases and tend to be stable after approximately 10-12 iterations. These experiments can verify the convergence of KMSA. (This figure is best viewed in color.)}
	\label{fig3}
\end{figure*}

Calculating the derivative of $\mathcal{L}\left(\bm{\alpha},\lambda\right) $ with respect to $\alpha_v$ and setting $\lambda$ to zero, we obtain

\begin{equation}
\label{eq12}
\left\{ {\frac{\partial \mathcal{L}\left(\bm{\alpha},\lambda\right)}{\partial \alpha_v} = \frac{\partial \mathcal{G}\left(\bm{\alpha},\bm{U}^1,\bm{U}^2, \cdots,\bm{U}^m\right)}{\partial \alpha_v} -\lambda = 0, v=1,2,\cdots,m  \atop  \frac{\partial \mathcal{L}\left(\bm{\alpha},\lambda\right)}{\partial \lambda} = \sum\limits_{v=1}^m \alpha_v-1 =0, ~~~~~~~~~~~~~~~~~~~~~~~~~~~~~~~~~~}\right.
\end{equation}

where
\begin{equation}
\label{eq13}
\begin{array}{l}
\frac{\partial \mathcal{G}\left(\bm{\alpha},\bm{U}^1,\bm{U}^2, \cdots,\bm{U}^m\right)}{\partial \alpha_v}  = \\
~~~~~~~r\alpha_v^{r-1} \left\{ tr\left\{ {\left( {\bm{U}^v} \right)^T\bm{K}^v\bm{P}^v\bm{K}^v\bm{U}^v} \right\} + r\kappa \right.\\
~~~~~~~+ \left. \sum\limits_{v\neq w} \frac{1}{2\eta} tr\left\{ {\left( {\bm{U}^v} \right)^T\bm{U}^v\left( {\bm{U}^w } \right)^T\bm{U}^w } \right\} \right\}.\\
\end{array}
\end{equation}

Because $r\kappa = \frac{r\kappa}{N}tr\left(\bm{I}\right)$, we can further transform $\frac{\partial \mathcal{G}\left(\bm{\alpha},\bm{U}^1,\bm{U}^2, \cdots,\bm{U}^m\right)}{\partial \alpha_v} $ as
\begin{equation}
\label{eq14}
\begin{array}{l}
\frac{\partial \mathcal{G}\left(\bm{\alpha},\bm{U}^1,\bm{U}^2, \cdots,\bm{U}^m\right)}{\partial \alpha_v}  = r\alpha_v^{r-1} tr\left( {\left( {\bm{U}^v} \right)^T\mathcal{J}^v \bm{U}^v} \right), \\
\end{array}
\end{equation}

where $\mathcal{J}^v = \bm{K}^v\bm{P}^v\bm{K}^v+ \frac{r\kappa}{N}\bm{I}+\sum\limits_{v\neq w} \frac{1}{2\eta}\bm{U}^w\left( {\bm{U}^w } \right)^T$. Therefore, we can obtain $\alpha_v$ as
\begin{equation}
\label{eq15}
\alpha_v = \frac{\left(1/ tr\left( {\left( {\bm{U}^v} \right)^T\mathcal{J}^v \bm{U}^v} \right) \right)^{1/\left(r-1\right)}}{\sum\limits_{v=1}^m\left(1/ tr\left( {\left( {\bm{U}^v} \right)^T\mathcal{J}^v \bm{U}^v} \right) \right) ^{1/\left(r-1\right)}}.
\end{equation}

It is notable that the value of $r$ ($r>1$) can directly influence the weighting factor $\alpha_v$. We analyze the influence as follows:

\begin{itemize}
	\item If $r$ infinitely approaches $1$, there is only one nonzero element $\alpha_i$, and $tr\left(\left( {\bm{U}^i} \right)^T\mathcal{J}^i \bm{U}^i\right)$ is the smallest among all $m$ views.
	\item  Conversely, if $r$ is infinite, all elements in $\bm{\alpha}$ tend to be equal to $1/m$.
\end{itemize}

After $\bm{\alpha},\bm{U}^1,\bm{U}^2, \cdots,\bm{U}^m$ are obtained, the low-dimensional representations for the $v$th view can be calculated as \ref{eq16}:

\begin{equation}
\label{eq16}
\small
\begin{array}{l}
\bm{Y}^v = \left[ {\bm{y}_1^v ,\bm{y}_2^v, \cdots ,\bm{y}_N^v} \right] = \left( {\bm{U}^v\bm{X}_\varphi ^v} \right)^T\bm{X}_\varphi ^v = \left( {\bm{U}^v} \right)^T\bm{K}^v.
\end{array}
\end{equation}

\subsection{Convergence of KMSA}

Because KMSA is solved by the alternating optimization strategy, it is essential to analyze its convergence.

\textbf{Theorem 1.} \textit{The objective function $\mathcal{G}\left(\bm{\alpha},\bm{U}^1,\bm{U}^2, \cdots,\bm{U}^m\right)$ in Eq. \ref{eq8} is bounded. The proposed optimization algorithm monotonically decreases the value of $\mathcal{G}\left(\bm{\alpha},\bm{U}^1,\bm{U}^2, \cdots,\bm{U}^m\right)$ in each step.}

\textbf{Lower Bound:} It is easy to see that there must exist one view (assumed as the $e$th view) that can make $\mathcal{K}_e= \alpha_e^r tr\left\{ {\left( {\bm{U}^e} \right)^T\bm{K}^e\bm{P}^e\bm{K}^e\bm{U}^e} \right\}$ the smallest among all views. Furthermore, there must exist two views (the $b$th and $c$th views) that can make $\mathcal{O}_{b,c} =\left(\alpha_b^r+\alpha_c^r \right) tr\left\{ {\left( {\bm{U}^b} \right)^T\bm{U}^b\left( {\bm{U}^c } \right)^T\bm{U}^c } \right\}$ the largest among all pairs of views. Because $\|\bm{\alpha}\|_r^r>0$, $\mathcal{G}\left(\bm{\alpha},\bm{U}^1,\bm{U}^2, \cdots,\bm{U}^m\right) \ge m\mathcal{K}_e +  \frac{C_m^2}{2\eta}\mathcal{O}_{b,c}$ can be proved. Therefore, $\mathcal{G}\left(\bm{\alpha},\bm{U}^1,\bm{U}^2, \cdots,\bm{U}^m\right)$ has a lower bound.

\begin{figure*}[htp]
	\centerline{\subfloat[Some image from Corel1K. There are 10 classes in this dataset, e.g., elephant, bus, dinosaur, flower, and horse. ]{\includegraphics[width=5in]{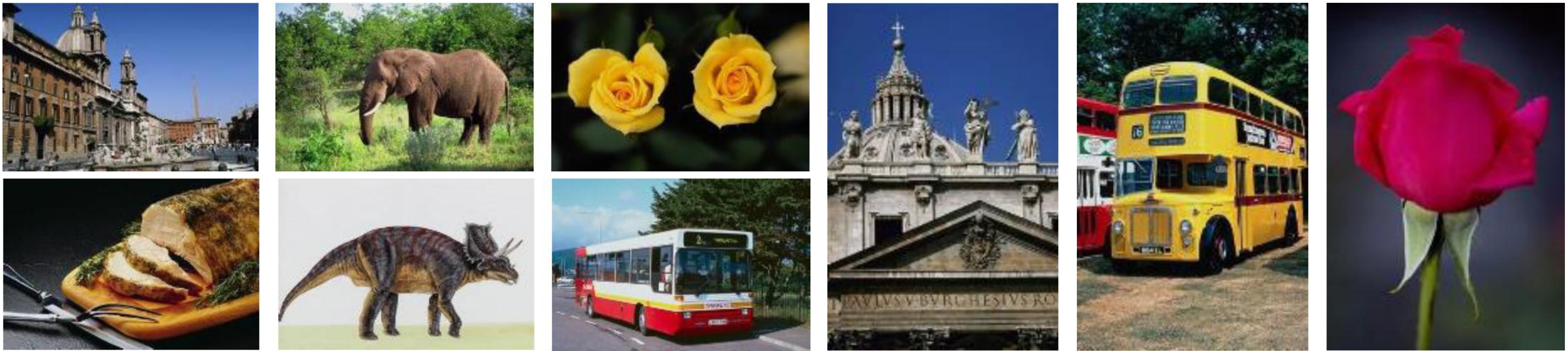}}}
	\centerline{\subfloat[Some image from Corel5K. Corel5K is an extension of Corel1K. It consists of 50 classes in total.]{\includegraphics[width=5in]{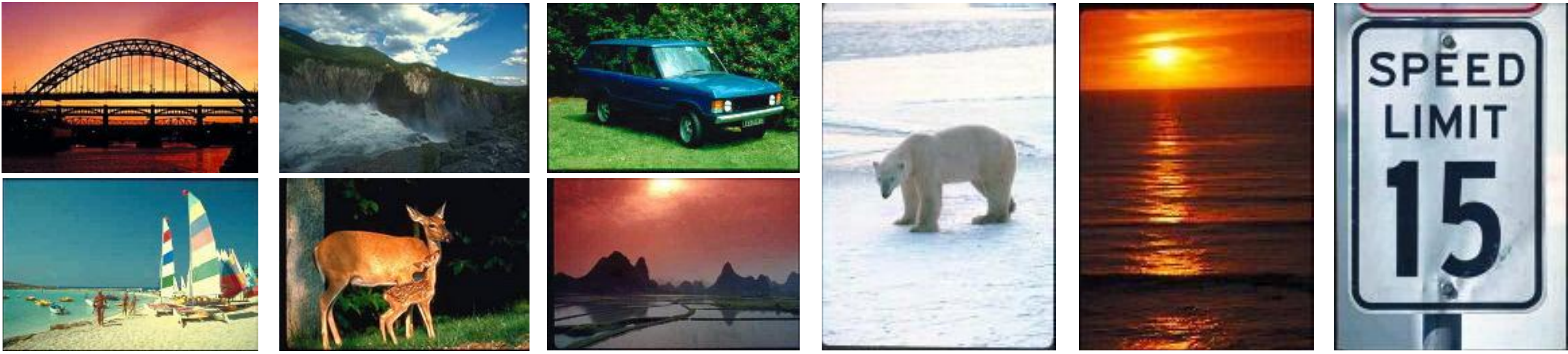}}}
	\centerline{\subfloat[Some image from Caltech101. Caltech101 is an image dataset that contains 101 classes and 1 background class, including faces, piano, football, airport, elephant, etc.   ]{\includegraphics[width=5in]{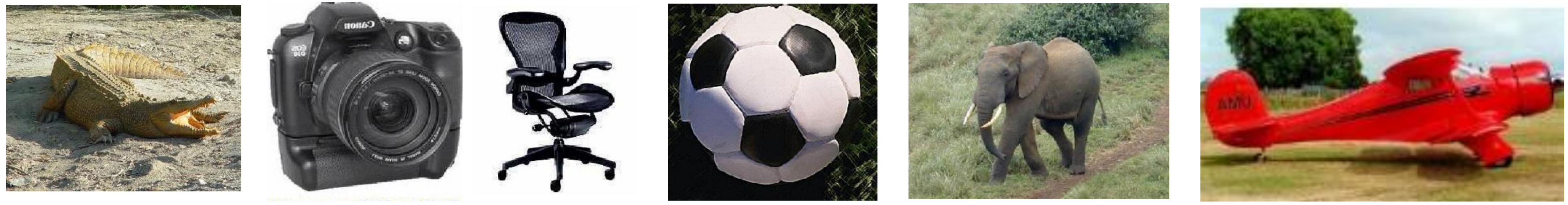}}}
	\caption{Some images from the Corel1K, Corel5K and Caltech101 datasets. (This figure is best viewed in color.)}
	\label{fig4}
\end{figure*}

\textbf{Monotone Decreasing:} During the optimization process, eigenvalue decomposition is adopted to solve $\bm{U}^1,\bm{U}^2, \cdots,\bm{U}^m $. Assume that $\bm{U}_t^v, t=1,2,\cdots,m,$ is calculated after the $t$-th main iterations. Because the solving method is based on eigenvalue decomposition, only the eigenvectors that correspond to the smallest $d$-th eigenvalues are maintained in $\bm{U}^v$. Therefore, in the process of updating $\bm{U}^v$ during the $(t+1)$-th main iteration, it is always true that

\begin{equation}
\begin{array}{l}
\mathcal{G}^{t+1}\left(\bm{\alpha}_t,\bm{U}_t^1,\bm{U}_t^2, \cdots,\bm{U}_{t+1}^v,\cdots,\bm{U}_t^m\right)\\
= \mathcal{A}tr(\bm{U}_{t+1}^v\left(\bm{U}_{t+1}^v\right)^T) = \mathcal{A} \sum\limits_{i=1}^d{\sigma_i^2} \leq r tr(\bm{U}_t^v\left(\bm{U}_t^v\right)^T) \\
= \mathcal{G}^t\left(\bm{\alpha}_t,\bm{U}_t^1,\bm{U}_t^2, \cdots,\bm{U}_{t}^v,\cdots,\bm{U}_t^m\right),
\end{array}
\end{equation}

where $\mathcal{A}$ is a constant because all the other variables remain unchanged. $\sigma_1,\sigma_2,\cdots,\sigma_d$ are the smallest $d$ eigenvalues of $\bm{U}_t^v$. Furthermore, the method for solving $\bm{\alpha}_t$ adopts the gradient descent, which always updates $\bm{\alpha}_t$ to make $\mathcal{G}^t$ smaller.

\textbf{Convergence Explanation:} Denote the value of $\mathcal{G}\left(\bm{\alpha},\bm{U}^1,\bm{U}^2, \cdots,\bm{U}^m\right)$ as $\mathcal{G}$, and let $\{\mathcal{G}^t\}_{t=1}$ be a sequence generated by the $t$-th main iteration of the proposed optimization. In addition, $\{\mathcal{G}^t\}_{t=1}$ is a bounded below monotone decreasing sequence based on the above theorem. Therefore, according to the bounded monotone convergence theorem \cite{rudin1976principles}, which asserts the convergence of every bounded monotone sequence, the proposed optimization algorithm converges.

Moreover, to further show the convergence of KMSA, we provide a figure to give the objective function values with the iterations. We extend LDA and PCA to multiview mode using KMSA and name them KMSA-LDA and KMSA-PCA. We record the objective function values with the corresponding numbers of iterations for these 2 methods on the Corel1K, Caltech101 and ORL datasets as shown in Fig. \ref{fig3}.

It can be seen that the objective function values of both KMSA-LDA and KMSA-PCA decrease as the number of iterations increases. The objective function values tend to be stable after 10-12 iterations. This result verifies that KMSA converges once a sufficient number of iterations are finished.

\subsection{Extension of Various DR Algorithms by KMSA}

The proposed KMSA can extend different dimension reduction algorithms to multiview mode. To facilitate the related research, we illustrate how to set $\bm{S}^v$ and $\bm{B}^v$ for different DR algorithms in the following:

\textbf{1. PCA:} $\bm{S}_{ij}^v = -1/N, i \neq j$, and $\bm{\mathcal{Q}}^v = \bm{I}$

\textbf{2. LPP:} $\bm{S}_{ij}^v = \exp\{-||x^v_i-x^v_j||^2/t  \}$ if $i\in N(j)$ or $j\in N(i)$ in the $v$th view, and $\bm{B}^v = \bm{D}^v$. $\bm{D}^v$ is a diagonal matrix, and $\bm{D}^v_{ii}$ is the sum of all elements in the $v$th line of $\bm{D}^v$.

\textbf{3. LDA:} $\bm{S}_{ij}^v = \delta_{l_i,l_j}/n_{l_i}$, and $ \bm{B}^v = \bm{I}- \frac{1}{N}\bm{ee}^T $, where $l_i$ is the label of the $i$th view. $n_{l_i}$ is the number of samples in the $i$th class. $\delta_{l_i,l_j} = - 1$ if $l_i \neq l_j$; otherwise, $\delta_{l_i,l_j} =  1$.

\textbf{4. SPP:} $\bm{S}^v = \bm{M}^v +\left(\bm{M}^v \right)^T + \left(\bm{M}^v \right)^T\bm{M}^v$. $\bm{B}^v= \bm{I}$ and $\bm{M}^v$ is constructed by sparse representation \cite{qiao2010sparsity}.

\section{Experiments}
\label{Experiment}

To verify the excellent performance of our proposed framework, we conduct several experiments on image retrieval (including the Corel1K \footnote{https://sites.google.com/site/dctresearch/Home/content-based-image-retrieval}, Corel5K and Holidays \footnote{http://lear.inrialpes.fr/~jegou/data.php} datasets) and image classification (including the Caltech101 \footnote{http://www.vision.caltech.edu/Image\_Datasets/Caltech101/Caltech101.html}, ORL \footnote{https://www.cl.cam.ac.uk/research/dtg/attarchive/facedatabase.html} and 3Sources \footnote{http://http://erdos.ucd.ie/datasets/3sources.html} datasets). In this section, we first introduce the details of the utilized datasets and methods for comparison in \ref{datasetsmethods}. Then, we present the experiments in \ref{ImageRetrieval} and \ref{ImageClassify}. The various experiments reveal the excellent performance of our proposed methods.

\begin{figure*}[htp]
	\centerline{
		\subfloat[Precision]{\includegraphics[width=1.8in]{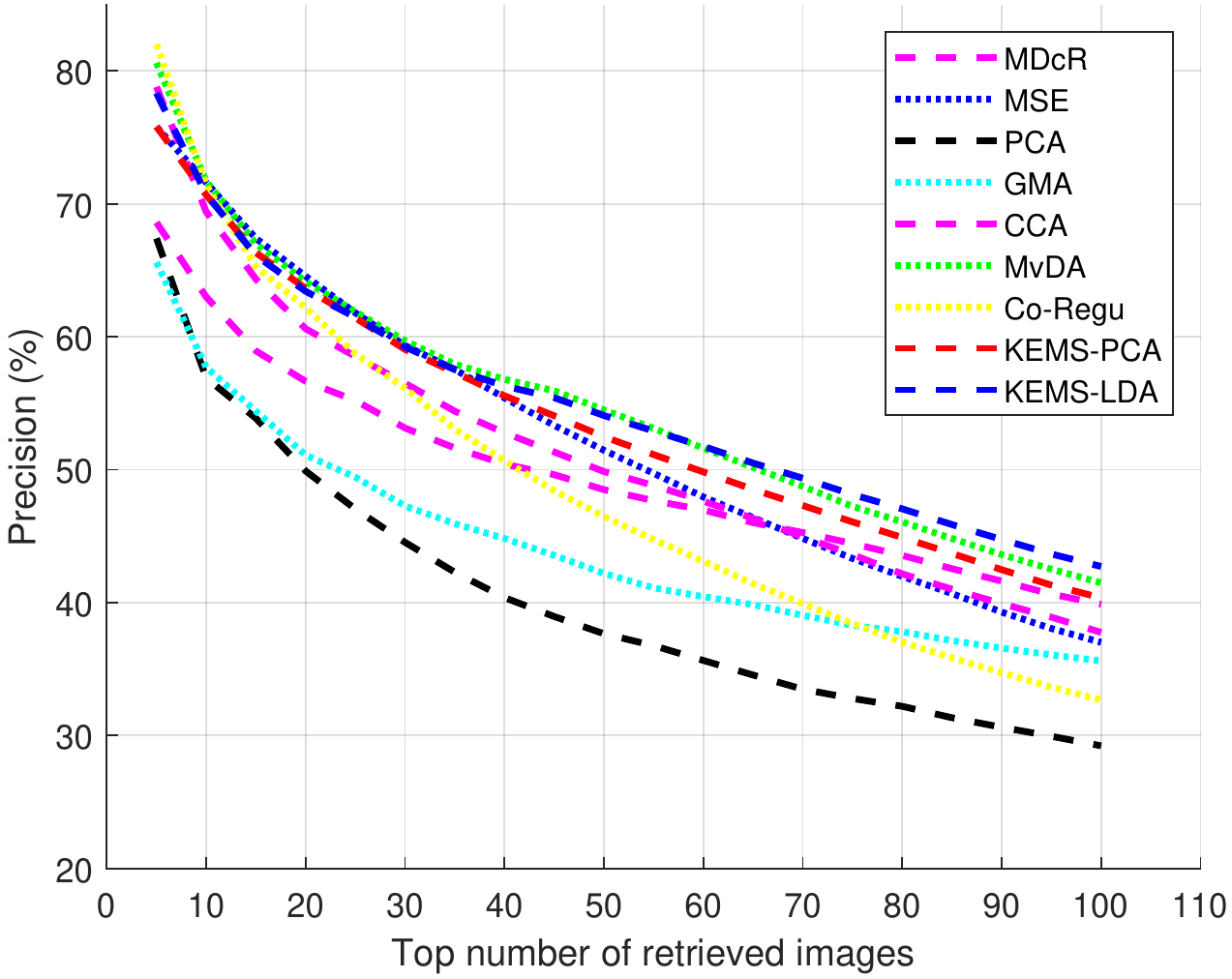}}
		\subfloat[Recall]{\includegraphics[width=1.8in]{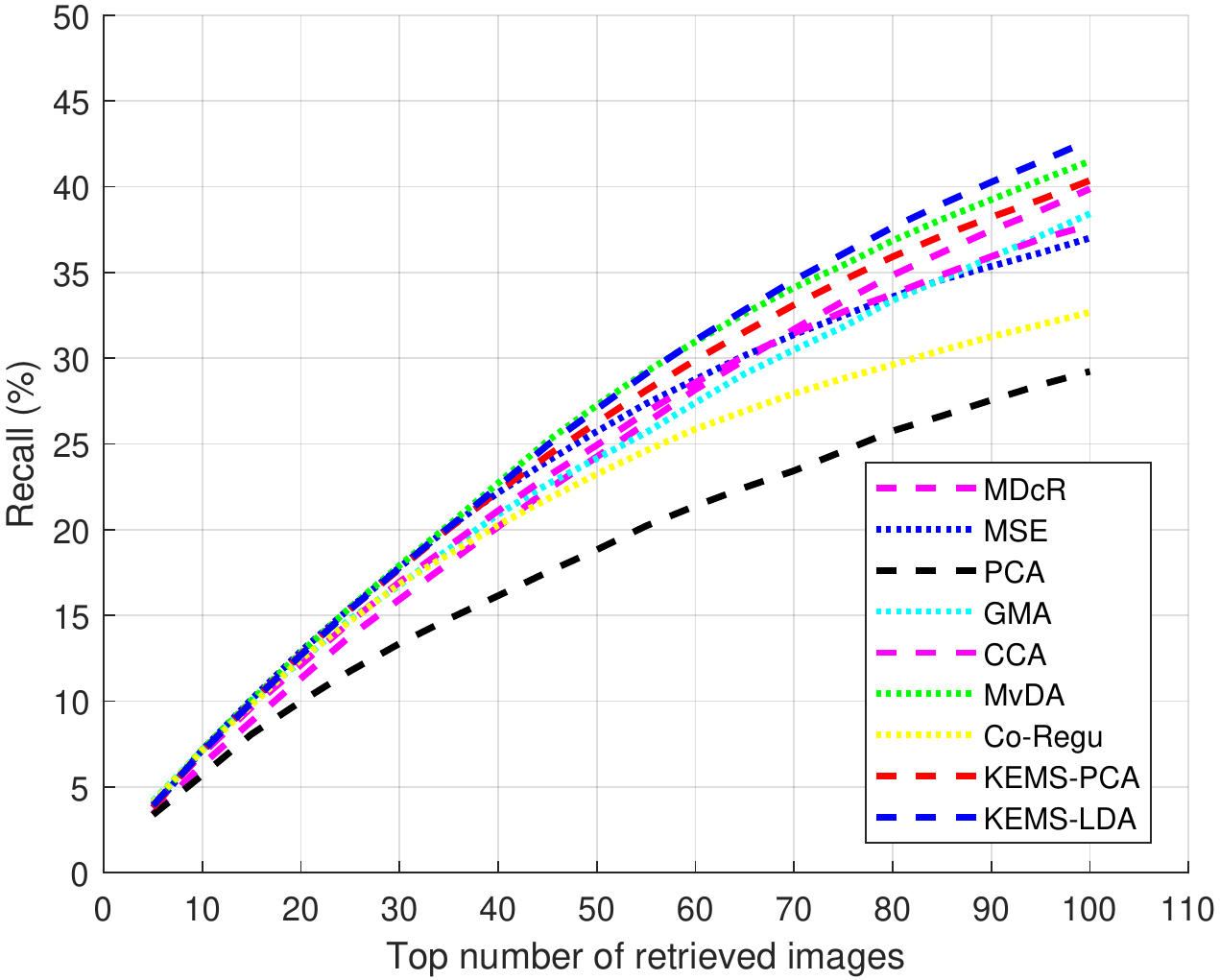}}
		\subfloat[PR-Curve]{\includegraphics[width=1.8in]{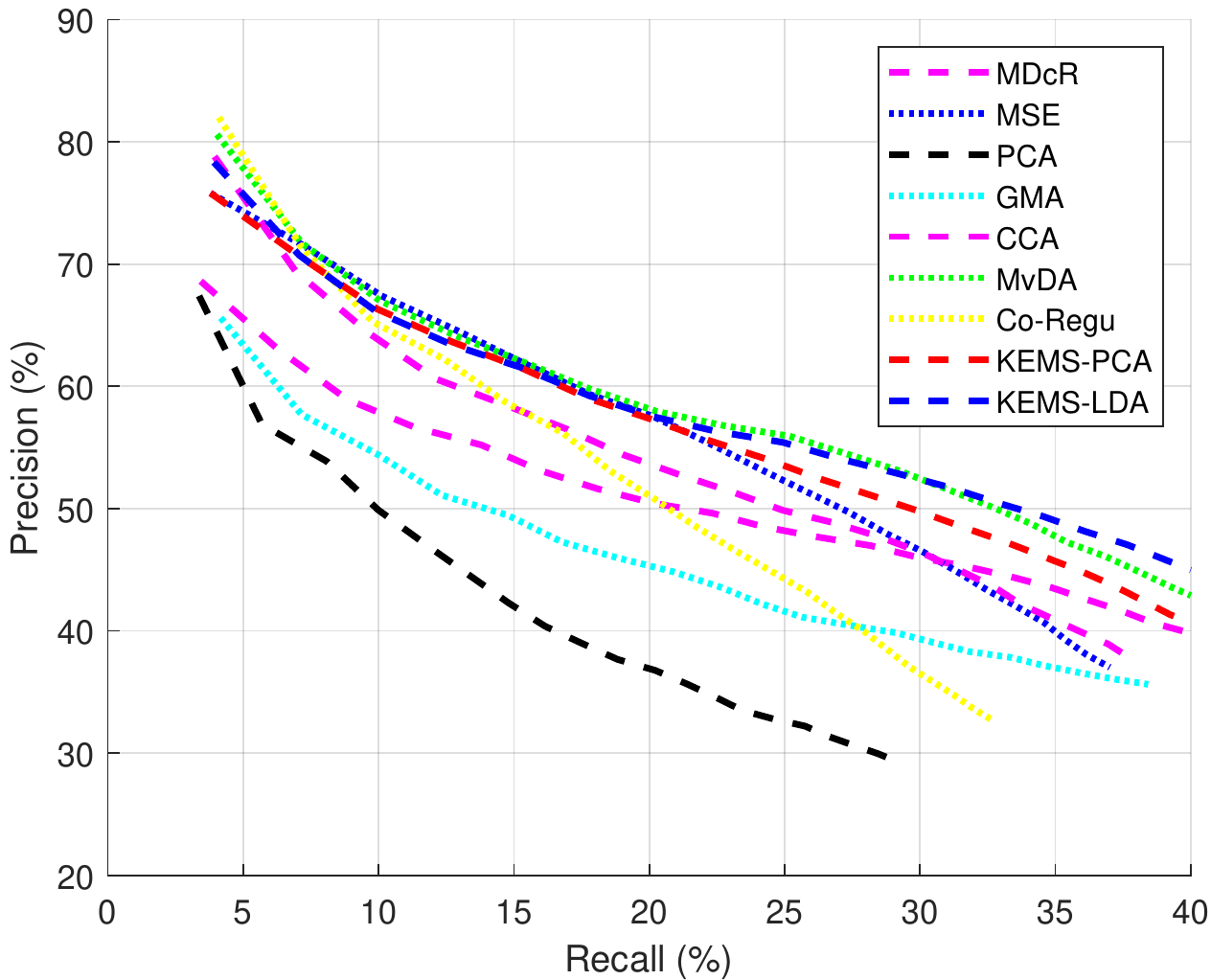}}
		\subfloat[F1-Measure]{\includegraphics[width=1.8in]{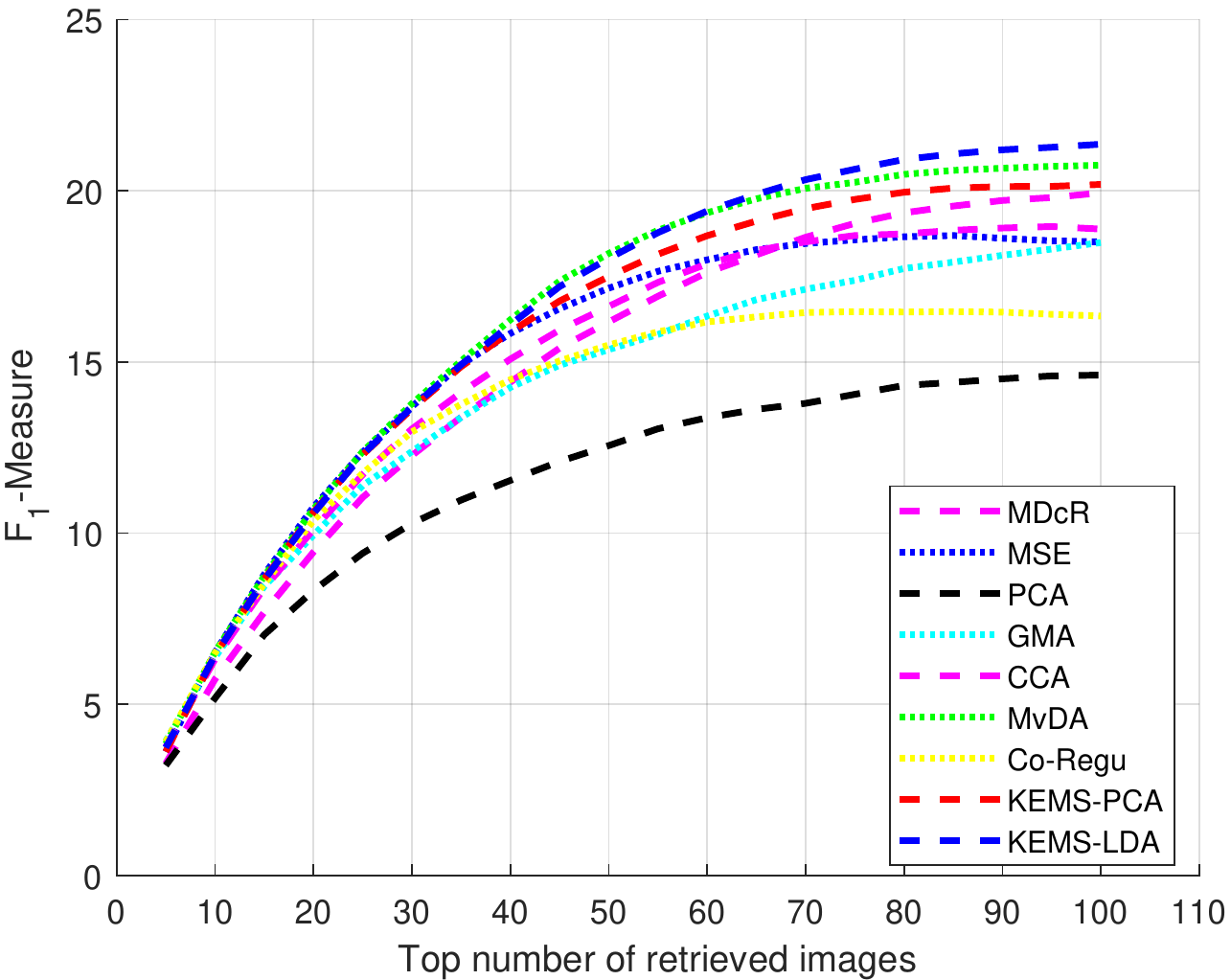}}}
	\caption{The average image retrieval results on Corel1K, where we repeated the experiments 20 times. KMSA-PCA outperforms all the other unsupervised multiview methods, and the performance of KMSA-LDA is the best in most situations. MvDA is a better method when the number of retrieved images is small. (This figure is best viewed in color.)}
	\label{fig5}
\end{figure*}

\begin{figure*}[htp]
	\centerline{
		\subfloat[Precision]{\includegraphics[width=1.8in]{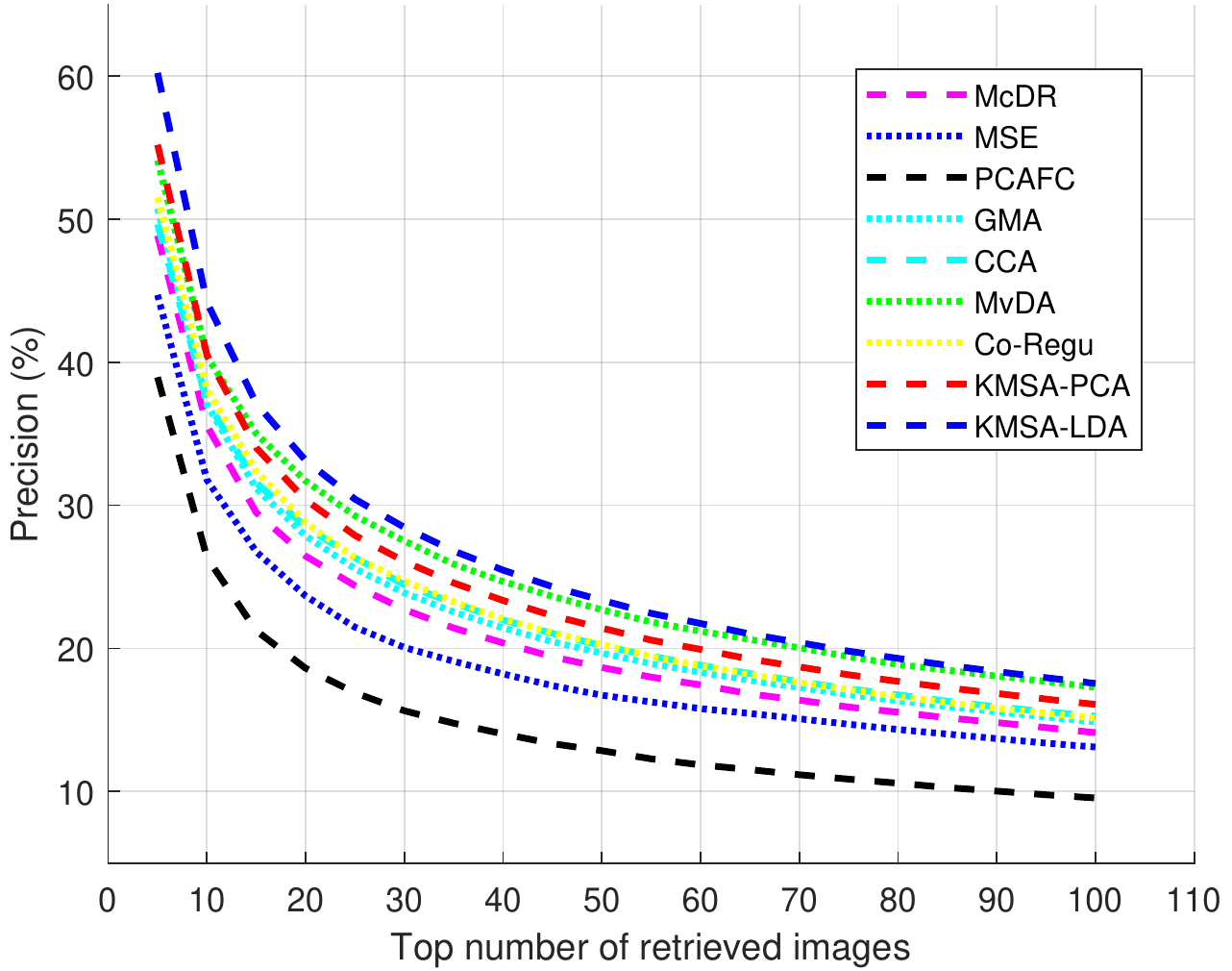}}
		\subfloat[Recall]{\includegraphics[width=1.8in]{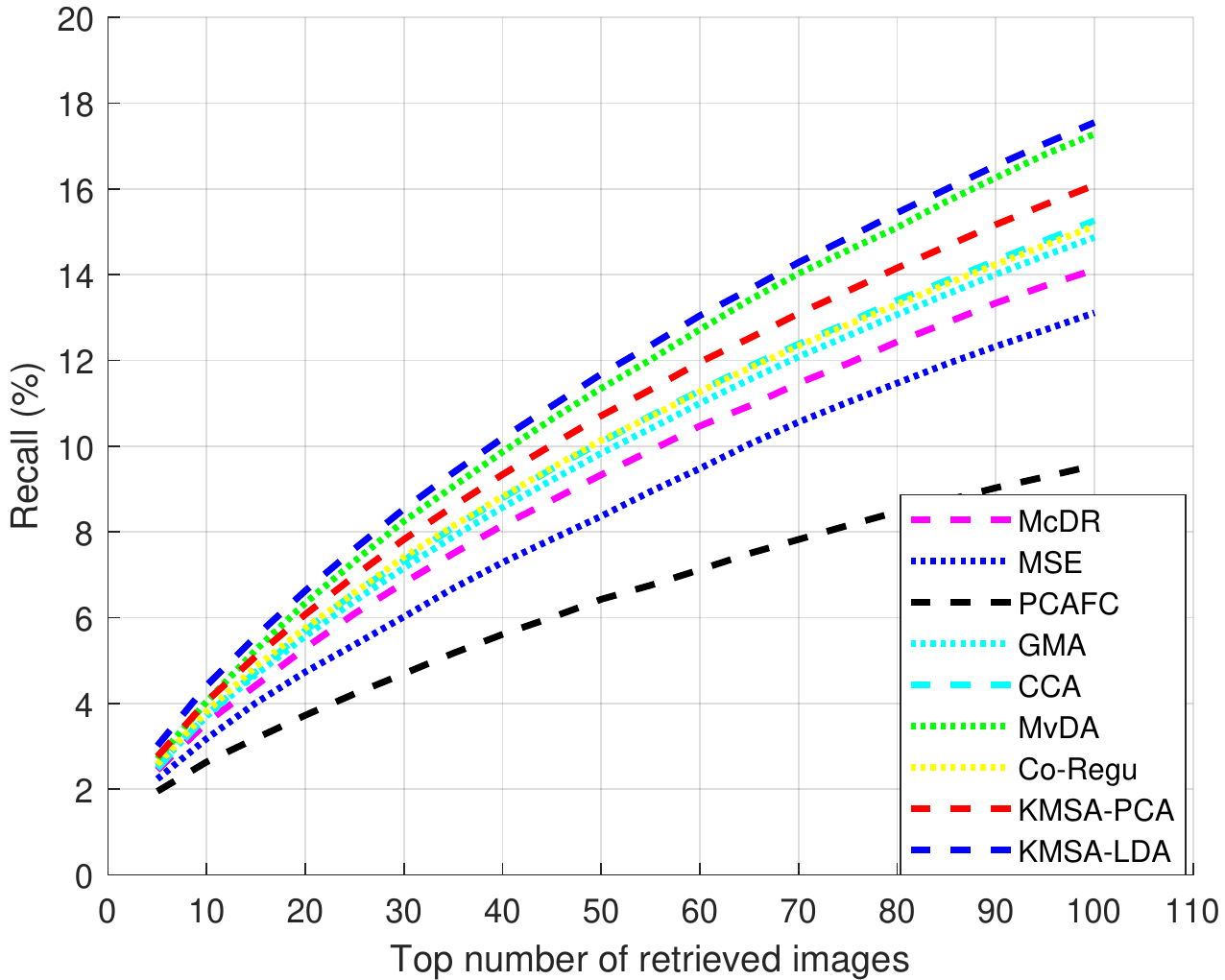}}
		\subfloat[PR-Curve]{\includegraphics[width=1.8in]{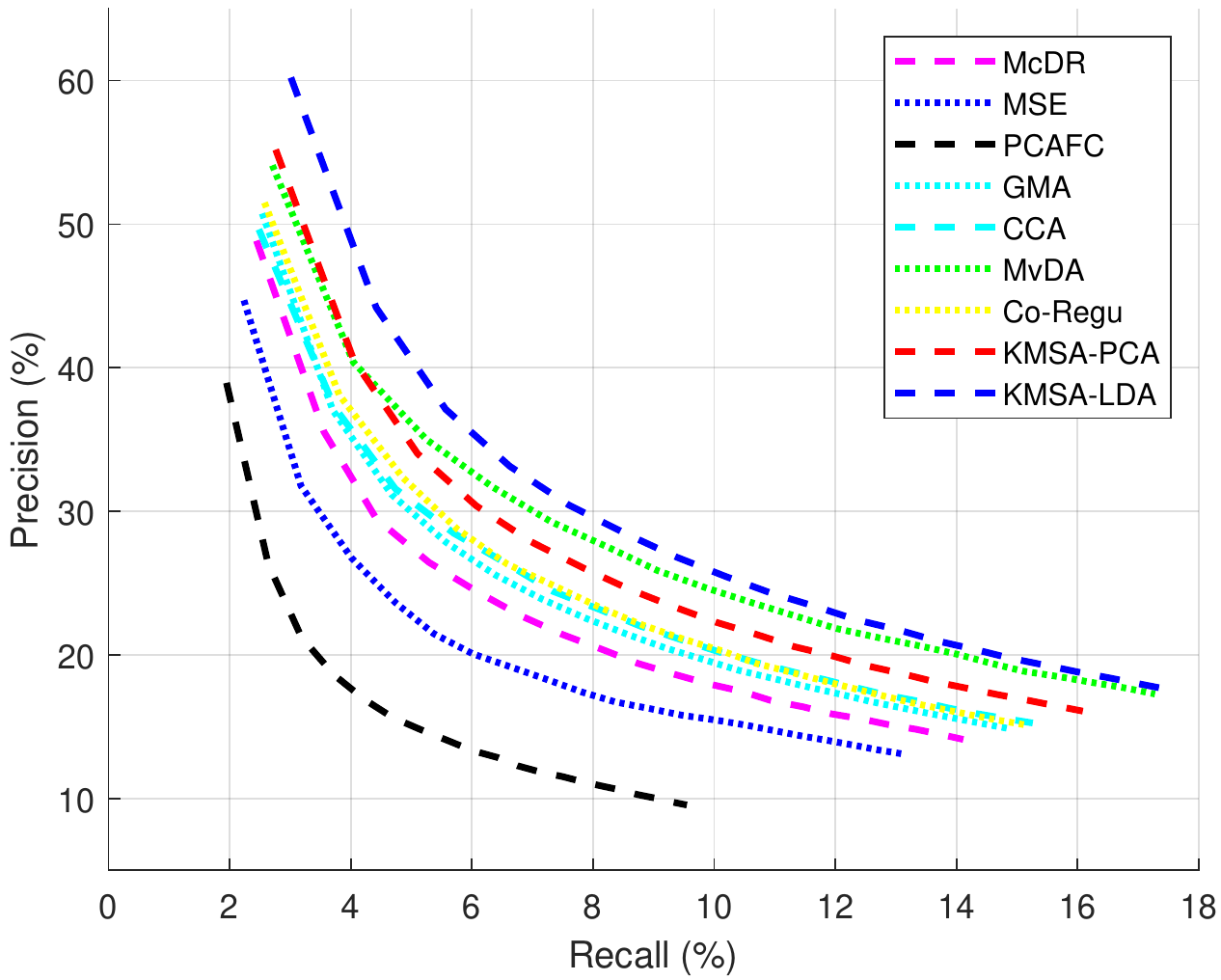}}
		\subfloat[F1-Measure]{\includegraphics[width=1.8in]{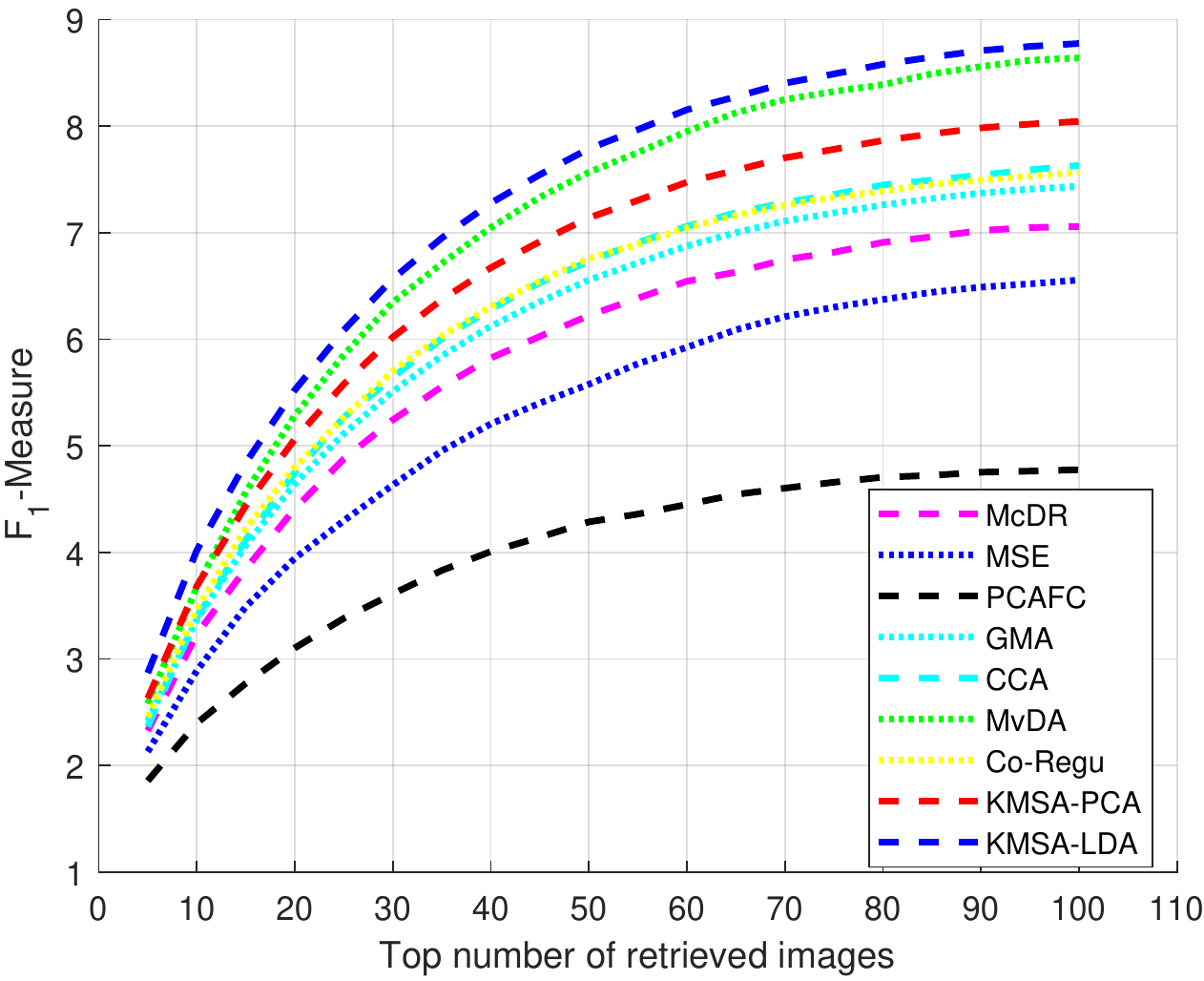}}}	
	\caption{The average image retrieval results on Corel5K, where we repeated the experiments 20 times. It is clear that KMSA can achieve the best performance in most situations. PCAFC performs worst because it is a single-view method. CCA and Co-Regu can also achieve ideal performances. (This figure is best viewed in color.)}
	\label{fig6}
\end{figure*}

\subsection{Datasets and Comparison Methods}
\label{datasetsmethods}

We introduce the utilized datasets and methods for comparison in this section. We conduct our experiments on image retrieval and multiview data classifications. The Corel1K, Corel5K and Holidays datasets are utilized for image retrieval, while Caltech101, ORL and 3Sources are utilized for multiview data classification. The details regarding the utilized datasets are as follows:

\textbf{Corel1K} is a specific image dataset for image retrieval. It contains 1000 images from 10 categories, e.g., bus, dinosaur, beach, and flower. There are 100 images in each category.

\textbf{Corel5K} is an extension of Corel1K for image retrieval. It contains 5000 images from 50 categories, including the images in Corel1K and some other images. Each category contains 100 images.

\textbf{Holidays} contains 1491 images corresponding to 500 categories,
which are mainly captured from various sceneries. The Holidays dataset is utilized for the image retrieval experiment.

\textbf{Caltech101} consists of 9145 images corresponding to 101 object categories and one background one. It is a benchmark image dataset for image classification.

\textbf{ORL} is a face dataset for classification. It consists of 400 faces corresponding to 40 people. Each person has 10 face images captured under different situations.

\textbf{3Sources} was collected from 3 well-known online news sources: BBC, Reuters
and the Guardian. Each source was treated as one view. 3Sources consists of 169 news articles in total.

We summarize the information of all views for these datasets in Table \ref{tab3}. In our experiment, we utilize several famous multiview subspace learning algorithms as comparison methods, including MDcR \cite{zhang2017flexible}, MSE \cite{Xia2010}, PCAFC \cite{jolliffe2011principal}, GMA \cite{sharma2012generalized}, CCA \cite{michaeli2016nonparametric} and MvDA \cite{kan2016multi}. It should be noted that GMA can also extend some DR methods into multiview mode. In this paper, we utilize GMA to represent the multiview extension of PCA. Meanwhile, PCAFC concatenates multiview data into one vector and utilizes PCA to obtain the low-dimensional representation. For KMSA, we set $\kappa = 0.1$, $\eta = -1$ and $r=3$ in our experiments. We adopt the Gaussian kernel for KMSA to extend multiview data into kernel spaces in our experiment.

\begin{table}[htbp]
	\center
	\caption{The information of all views for these datasets. The utilized features include the microstructure descriptor (MSD) \cite{liu2011image}, Gist \cite{oliva2001modeling}, the histograms of oriented gradients (HOG) \cite{dalal2005histograms}, the grayscale intensity (GSI), local binary patterns (LBPs) \cite{Ojala2002}, and the edge direction histogram (EDH) \cite{gao2008image}. BBC, Reuters and Guardian are 3 well-known online news sources, which are utilized as 3 views. }
	\begin{tabular}
		{cccc}
		\Xhline{1.2pt}
		Dataset &  View 1  & View 2  & View 3 \\
		
		\hline
		\hline
		Corel1K	&  MSD  & Gist  & HOG  \\
		
		Corel5K &MSD & Gist  & HOG  \\
		
		Holidays  &MSD  & Gist  & HOG   \\
		
		Caltech101  &MSD & Gist  & HOG  \\
		
		ORL&  GSI & LBP   &  EDH \\
		
		3Sources& BBC & Reuters & Guardian  \\
		
		\Xhline{1.2pt}
	\end{tabular}
	\label{tab3}
\end{table}

\begin{table*}[htbp]
	\center
	\caption{The average values of the precision ($P\%$), recall ($R\%$), mAP ($\%$) and $F_1$-Measure of different methods on the Holidays dataset. We repeated the experiments 20 times. It is clear that KMSA-LDA and KMSA-PCA are the 2 best methods. MvDA and Co-Regu can also achieve ideal performances. PCAFC is the worst because it is a single-view method and thus cannot fully utilize the multiview data. }
	\begin{tabular}
		{ccccccccccc}
		\hline
		\diagbox[width=5.8em]{\tiny{Criteria}}{\tiny{Methods}}
		& MDcR & MSE & PCAFC & GMA & CCA & MvDA & Co-Regu & KMSA-PCA &KMSA-LDA\\
		\hline
		
		Precision &77.69& 77.48 & 62.84 & 77.91&77.07 &80.24 & 78.04&78.84 &\textbf{80.73} \\
		
		Recall   & 60.05& 59.81 & 48.49 & 60.14&59.36 &61.91 &60.06 &60.58 &\textbf{62.21}\\
		
		mAP      & 89.08 & 88.74 & 77.22 & 89.22 & 88.43& 90.02&88.92 &89.64 &\textbf{90.77}\\
		
		F1-Measure& 67.74 & 67.51 & 54.74 &67.88 &67.07 &69.89 & 67.88 &68.51 &\textbf{70.27} \\
		
		\hline
	\end{tabular}
	\label{tab4}
\end{table*}

\begin{figure*}[htp]
	\centerline{
		\subfloat[30$\%$ of the Samples are for Training]{\includegraphics[width=2.5in]{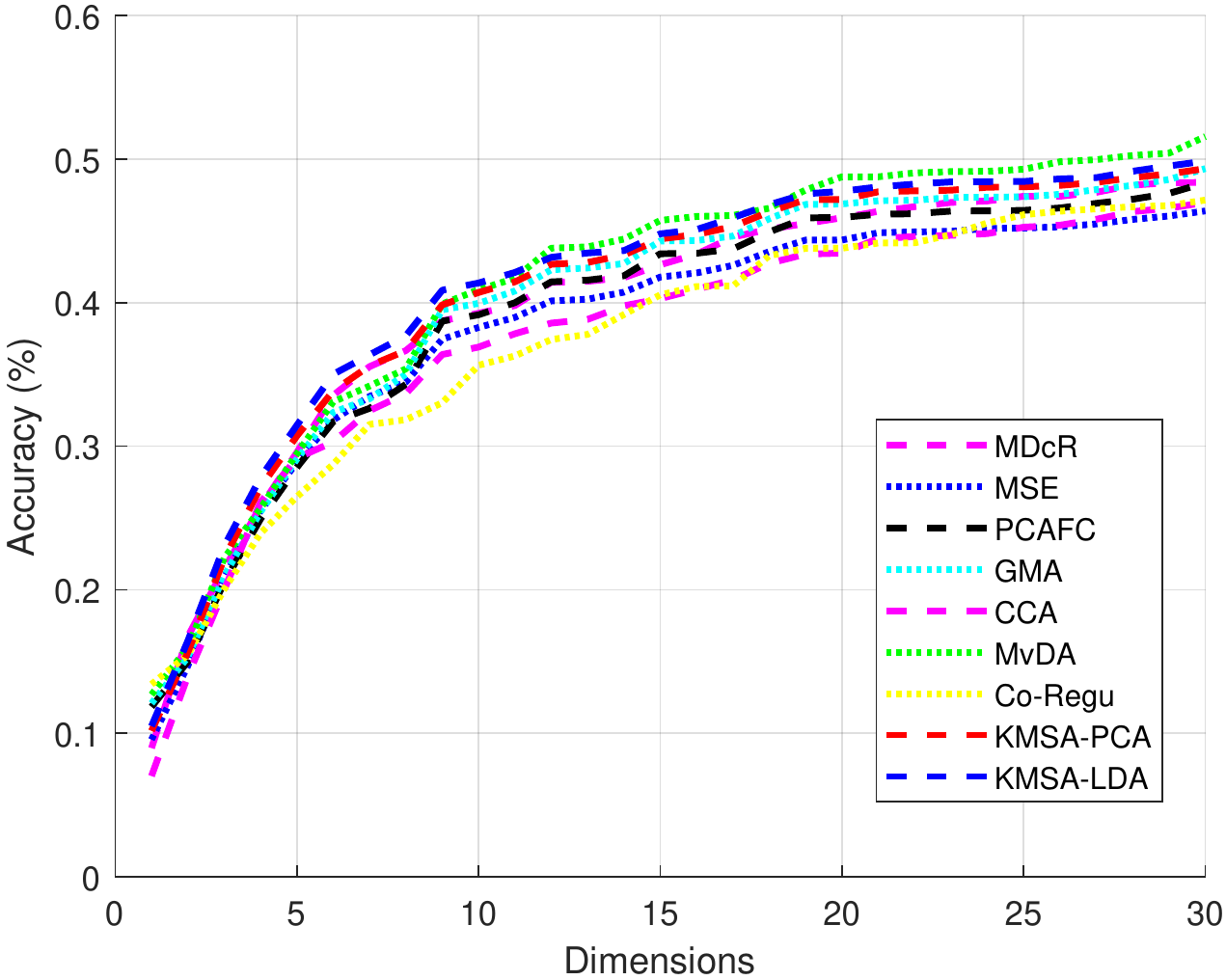}}
		~~~~~~~~\subfloat[50$\%$ of the Samples are for Training]{\includegraphics[width=2.5in]{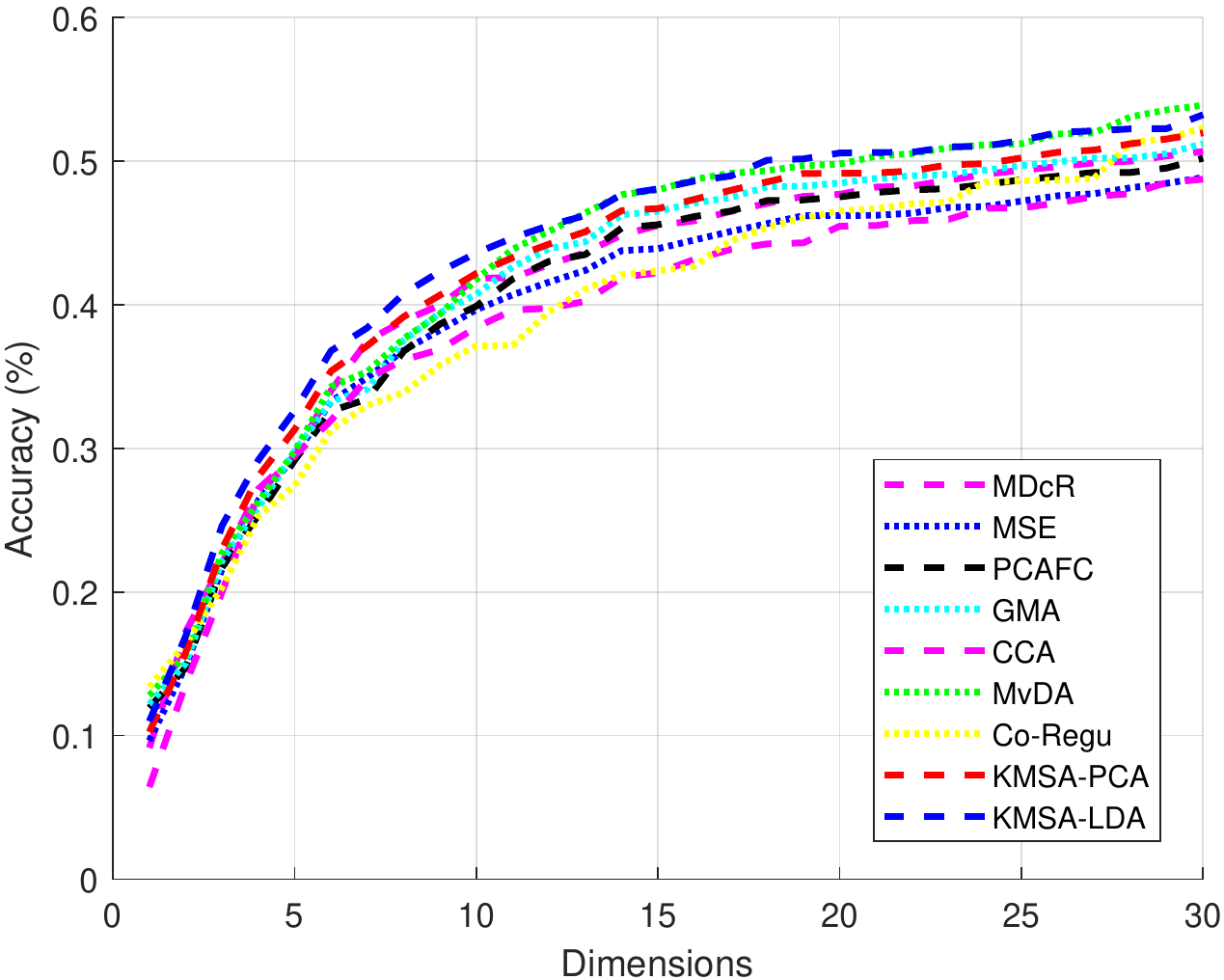}}}
	\caption{The average classification results on the Caltech101 dataset, where we repeat the experiments 20 times. These 2 figures randomly select different proportions of samples as the training ones. With the increase in dimensions, the performances of all methods improve. KMSA-LDA and KMSA-PCA are the 2 best methods in most situations. (This figure is best viewed in color.)}
	\label{fig7}
\end{figure*}

\subsection{Image Retrieval}
\label{ImageRetrieval}
In this section, we conduct experiments on the Corel1K, Corel5K and Holidays datasets for image retrieval.

For the Corel1K dataset, we randomly select 100 images as queries (each class has 10 images), while the other images are assigned as galleries. The MSD \cite{liu2011image}, Gist \cite{oliva2001modeling} and the HOG \cite{dalal2005histograms} are utilized to extract different features for multiple views. We utilize all methods to project multiview features into a 50-dimensional subspace and adopt the $l1$ distance for image retrieval. All experiments are conducted on the low-dimensional representations from the best view. We repeat the experiment 20 times and calculate the mean values of the Precision (P), Recall (R) and F1-Measure (F1). The results are shown in Fig. \ref{fig5}.

It is clear that KMSA-PCA can achieve a better performance than that of the other unsupervised multiview algorithms. Meanwhile, KMSA-LDA outperforms MvDA. It is shown that KMSA is an ideal framework for extending DR algorithms into the multiview case and achieves a better performance. Furthermore, even though PCAFC concatenates all views into one single vector, it cannot achieve a good performance because PCA is essentially a single-view method.

For the Corel5K dataset, we randomly select 500 images as queries (each class has 10 images), while the other images are assigned as galleries. The MSD \cite{liu2011image}, Gist \cite{oliva2001modeling} and
the HOG \cite{dalal2005histograms} are also utilized as the descriptors to extract features for multiple views. We utilize all methods to project multiview features into a 50-dimensional subspace and adopt the $l1$ distance to finish the task of image retrieval. The experimental settings are the same as as for the Corel1K dataset. The results are shown in Fig. \ref{fig6}.

As can been seen in Fig. \ref{fig6}, KMSA-LDA outperforms all the other methods in most situations. In addition, as an unsupervised method, the performance of KMSA-PCA is better. It is obvious that KMSA-LDA is a better method than KMSA-PCA. This is because label information can be fully considered by KMSA-LDA. Subspaces constructed by KMSA-LDA can better distinguish multiview data with different labels. Furthermore, MDcR and Co-Regu \cite{kumar2011co} are two good methods. PCAFC has the worst performance because it cannot fully exploit the information from the multiview data.

For the Holidays dataset, there are 3 images in one class. For each class, we randomly select 1 image as the query, with the other 2 images as the galleries. The MSD \cite{liu2011image}, Gist \cite{oliva2001modeling} and the HOG \cite{dalal2005histograms} are exploited to extract different features for multiple views. All methods are conducted to project multiview features into a 50-dimensional subspace. The experiments are conducted 20 times, and we calculate the mean values of those indices in Table \ref{tab4}:

\begin{table*}[htbp]
	\center
	\caption{The mean classification accuracies ($\%$) on the ORL dataset, where we repeat the experiments 20 times. It can be seen that KMSA-LDA is the best method and that KMSA-PCA outperforms the other unsupervised methods. MDcR and Co-Regu are not as good as the other methods. MvDA can also achieve an ideal performance.}
	\begin{tabular}
		{cccccccccccc}
		\hline
		Percentage& Dim
		& MDcR & MSE & PCAFC & GMA & CCA & MvDA & Co-Regu & KMSA-PCA &KMSA-LDA\\
		\hline
		
		\multirow{3}{*}{30$\%$}
		& 10 &58.10& 63.25 &   60.23 & 56.19 & 62.50 &64.52 & 60.48 & 64.16 & \textbf{67.42}  \\
		
		& 20&68.45 & 73.86 & 67.19 & 65.83& 72.26&  77.26& 67.86 & 74.56 &  \textbf{77.03}\\
		
		& 30&71.19 & 78.31 & 74.33 &70.83 & 77.26 & 84.20 & 74.52 &79.44 &\textbf{84.55}\\
		\hline
		\multirow{3}{*}{50$\%$}
		& 10 &68.69 & 70.22& 72.50 & 72.50& 72.83 &76.50  & 64.67 & 74.23 &\textbf{78.64}  \\
		
		& 20&79.44 & 81.58 & 79.83 &79.50 & 82.33&  87.17& 76.67 & 83.73 & \textbf{87.28} \\
		
		& 30&83.33 & 87.27 & 84.00 &83.67 & 85.83 & 90.17 &  80.50& 87.50 &\textbf{92.49}\\
		\hline
	\end{tabular}
	\label{tab5}
\end{table*}

From Table \ref{tab4}, we can also find that KMSA-PCA and KMSA-LDA can achieve the best performances in most situations. Co-Regu and MvDA can also obtain good results. Since PCAFC is a single-view method, it achieves the worst performance.

\begin{figure*}[htp]
	\centerline{
		\subfloat[30$\%$ of the Samples are for Training]{\includegraphics[width=3.5in]{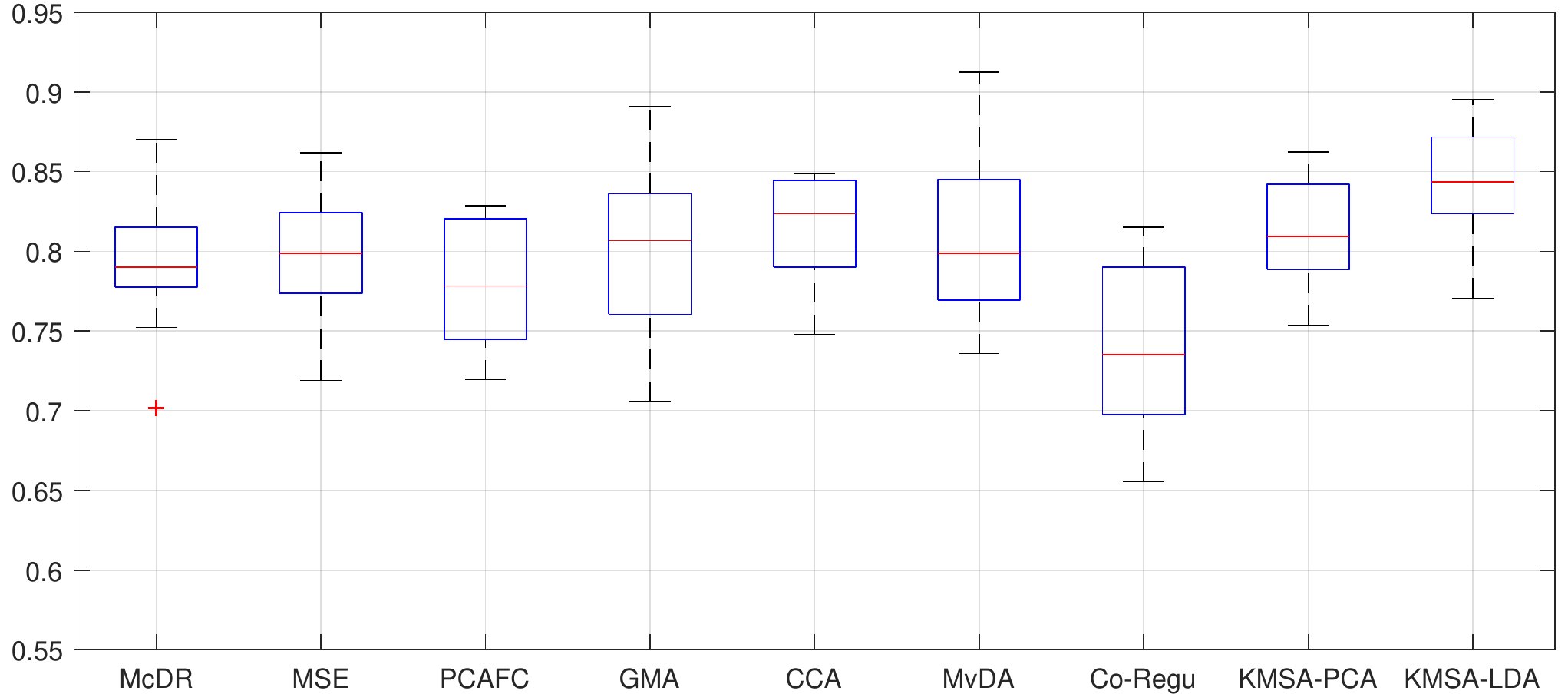}}	
		\subfloat[50$\%$ of the Samples are for Training]{\includegraphics[width=3.5in]{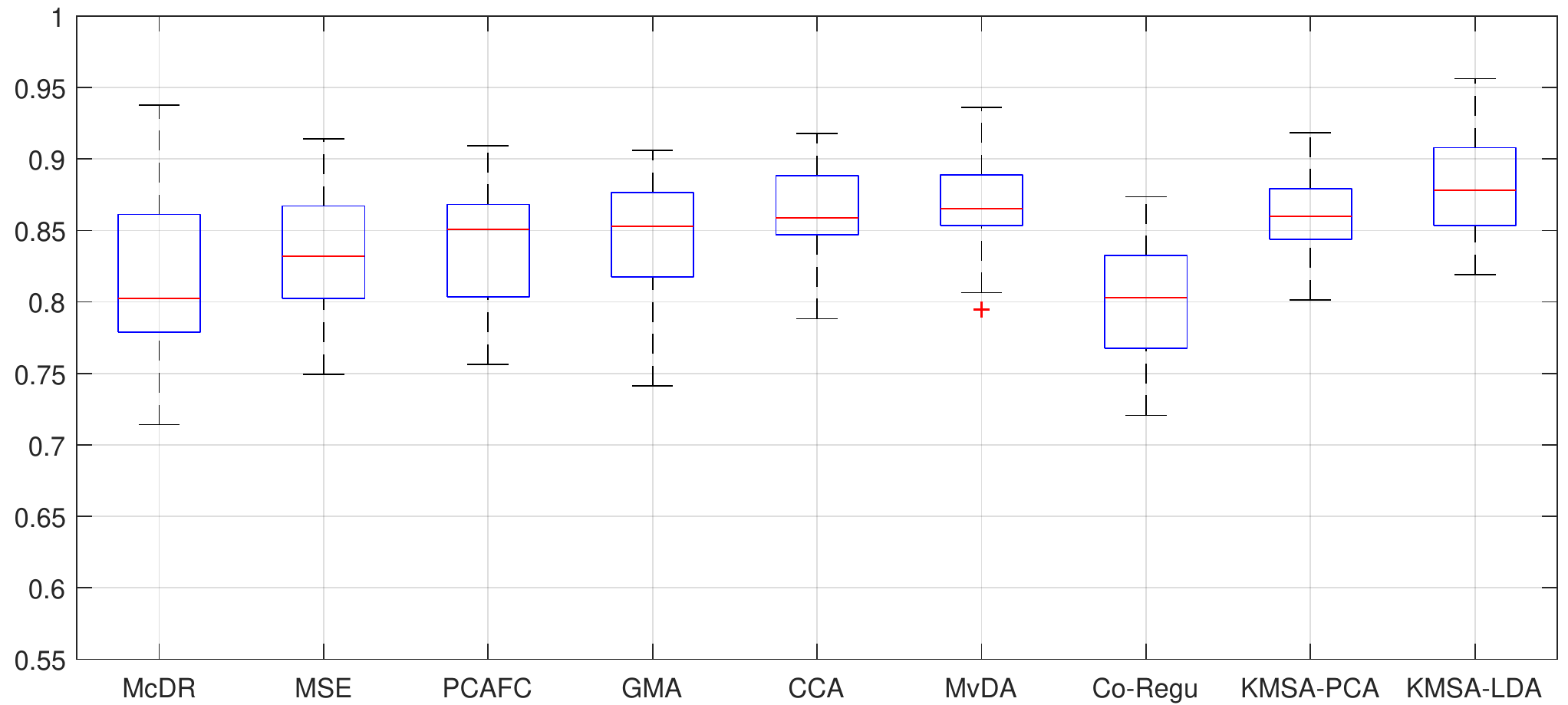}}}
	\caption{The average classification results on the 3Sources dataset, where we repeated the experiments 20 times. KMSA-LDA outperforms the other methods. Most of the multiview DR methods can achieve the ideal performance.}
	\label{fig8}
\end{figure*}

\subsection{Classification of Multiview Data}
\label{ImageClassify}

In this section, we conduct classification experiments on 3 datasets (including Caltech101, ORL and 3Sources) to verify the effectiveness of our proposed method.

For the Caltech101 dataset, we randomly select 30$\%$ and 50$\%$ of the samples as the training samples, with the other samples being assigned as the testing ones. The MSD \cite{liu2011image}, Gist \cite{oliva2001modeling} and the HOG \cite{dalal2005histograms} are utilized to extract different features for multiple views. All the methods are utilized to project multiview features into subspaces with different dimensions ($1-30$). 1NN is utilized to classify the testing samples. This experiment is conducted 20 times, and the mean results of all methods are shown in Fig. \ref{fig7}.

For the ORL dataset, we also randomly select 30$\%$ and 50$\%$ the samples as the training ones. The grayscale intensity, LBP \cite{Ojala2002} and EDH \cite{gao2008image} are utilized as the 3 views. The operations for this experiment are same as those on the Caltech101 dataset. 1NN is utilized as the classifier. We conduct this experiment 20 times, and the mean classification results for different dimensions can be found in Table \ref{tab5}.

It can be seen in Fig. \ref{fig7} and Table \ref{tab5} that with the increase in dimensions, the performances of all methods improve. KMSA-LDA is better than MvDA, while KMSA-PCA is the best unsupervised multiview method in our experiment. This is because KMSA can better exploit the information from the multiview data to learn the ideal subspaces, fully considering the multiple views and assigning reasonable weights to them automatically according to their importance.  Furthermore, KMSA adopts a co-regularized term to minimize the divergence between different views to help all views learn from each other. All these factors ensure that KMSA achieves a good performance. Moreover, it can be found that the performance of KMSA is very close to that of MvDA in some situations. Compared with those famous multiview methods with good performances, KMSA can
flexibly extend the single-view methods to their multiview modes and achieve a good performance in most situations. This is the starting point of the proposed KMSA.

For the 3Sources dataset, we also randomly select 30$\%$ and 50$\%$ of the samples as the training ones. It is a benchmark multiview dataset that consists of 3 views. We utilize all the methods to construct the 30-dimensional representations and adopt 1NN to classify the testing ones. The boxplot figures are shown in Fig. \ref{fig8}. All the experiments above verify the superior performance of KMSA. It can extend different DR methods to the multiview mode. From the experimental results, KMSA-LDA is better than MvDA, and KMSA-PCA outperforms the other unsupervised methods in most situations.

\subsection{Discussions on KMSA}

It is essential to discuss the factors influencing the performance of KMSA. We conducted experiments to show the influence of the self-weighting and kernelization schemes. We fix all weights to $1/m$ and show the performances of KMSA-PCA and KMSA-LDA on the Holidays dataset in Table \ref{tab6}. Furthermore, we compare KMSA with some kernelized versions of the DR methods, including kernel canonical correlation analysis (KCCA) \cite{lisanti2017multichannel} and kernel principle component analysis (KPCA) \cite{fan2019exactly}, to prove that the performance improvement of KMSA is not due to only the kernelization of single-view data. Similar to KMSA, we adopt the Gaussian kernel for KCCA and KPCA in this experiment. All the settings of the experiment in Table \ref{tab6} are the same as those for the Holidays dataset.

\begin{table}[htbp]
	\center
	\caption{The influence of the self-weighting and kernelization scheme for KMSA on the Holidays dataset. ``Fixed ($1/m$)'' means that all the weights of multiple views are fixed as $1/m$, and ``Flexible'' means that all the weights are learned automatically. ``Kernelized'' means that the corresponding methods are kernelized versions.  }
	\begin{tabular}
		{ccccccc}
		\hline
		\diagbox[width=5.8em]{\tiny{Methods}}{\tiny{Criteria}}
		&  & Precision & Recall & mAP & F1-Measure \\
		\hline
		
		\multirow{2}*{KMSA-PCA} & Fixed ($1/m$)  & 78.34  & 60.23 & 88.55 & 68.10  \\
		~ & Flexible & 78.84 & 60.58 & 88.92 &68.51 \\
		\hline
		
		\multirow{2}*{KMSA-LDA} & Fixed ($1/m$)  & 80.22   & 62.11 &  89.89 &  70.01 \\
		~ & Flexible & \textbf{80.73}& \textbf{62.21}& \textbf{90.77} & \textbf{70.27} \\
		\hline
		\hline
		\multirow{2}*{Kernelized} & KCCA  & 78.69  & 60.31  &88.76 & 68.28  \\
		~&KPCA & 77.45& 59.12  &87.15 & 67.05  \\

		\hline
	\end{tabular}
	\label{tab6}
\end{table}

\begin{figure*}[htbp]
	\centerline{
		\subfloat[Original data]{\includegraphics[width=2.3in]{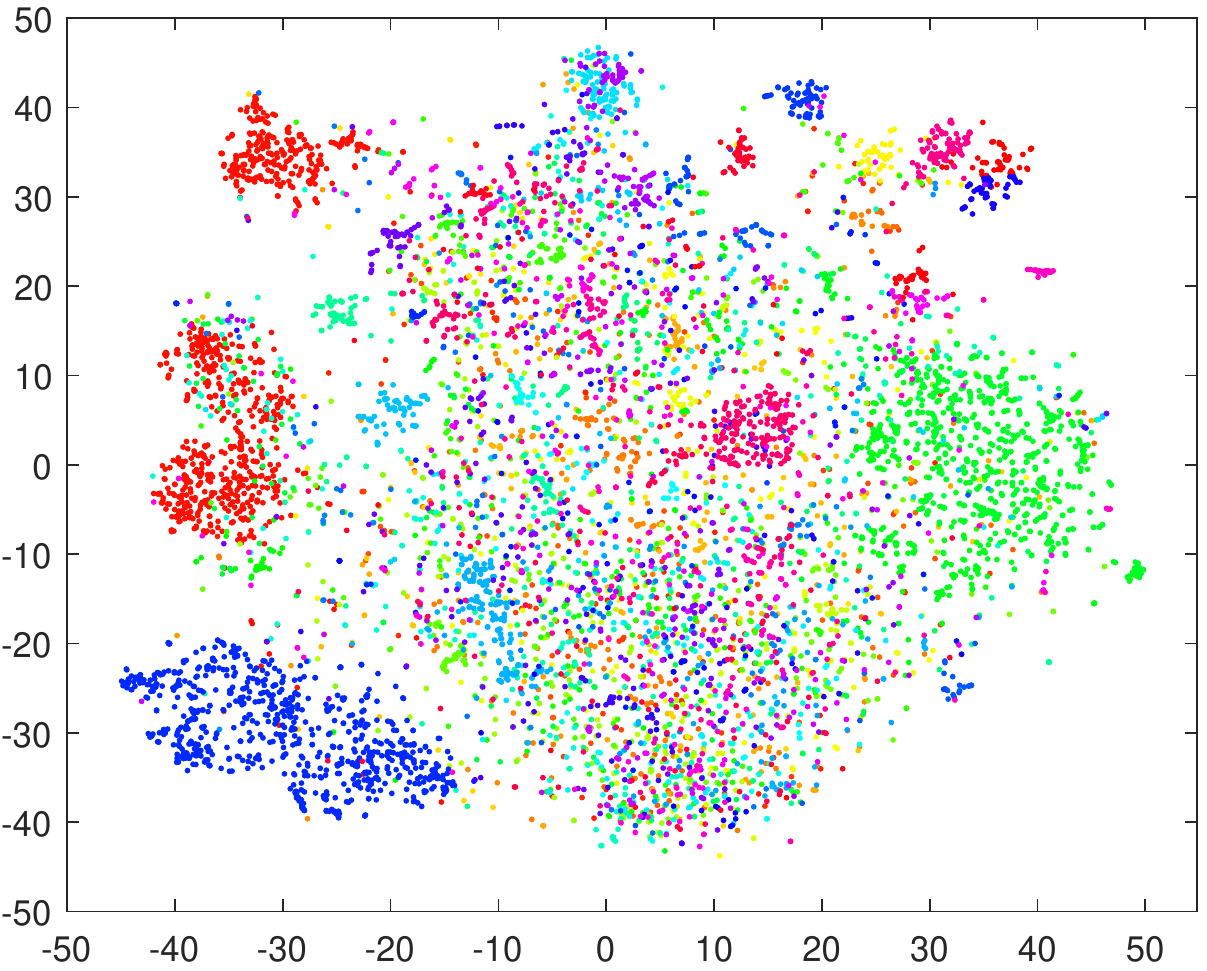}}	
		\subfloat[KMSA-PCA]{\includegraphics[width=2.3in]{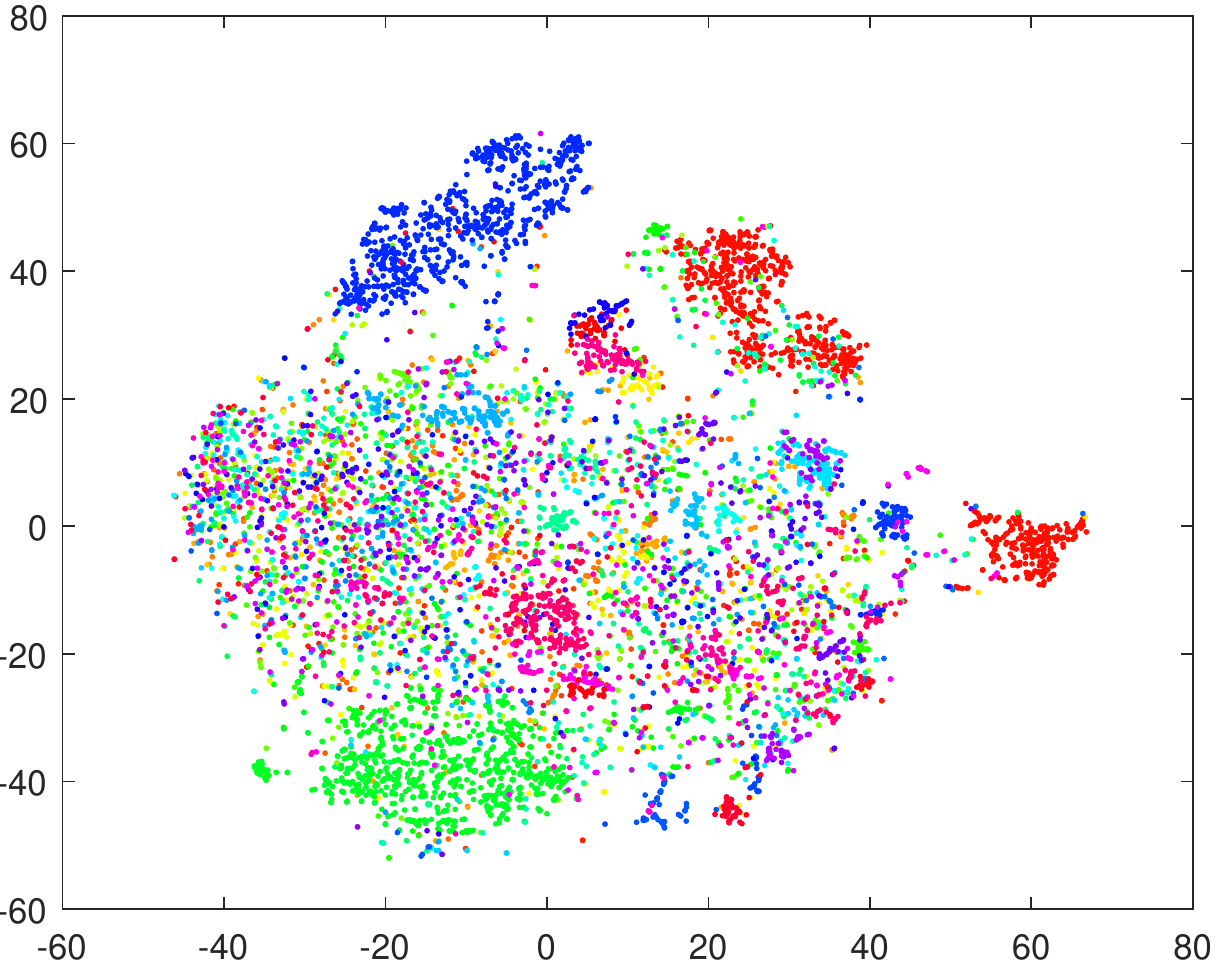}}
		\subfloat[KMSA-LDA]{\includegraphics[width=2.3in]{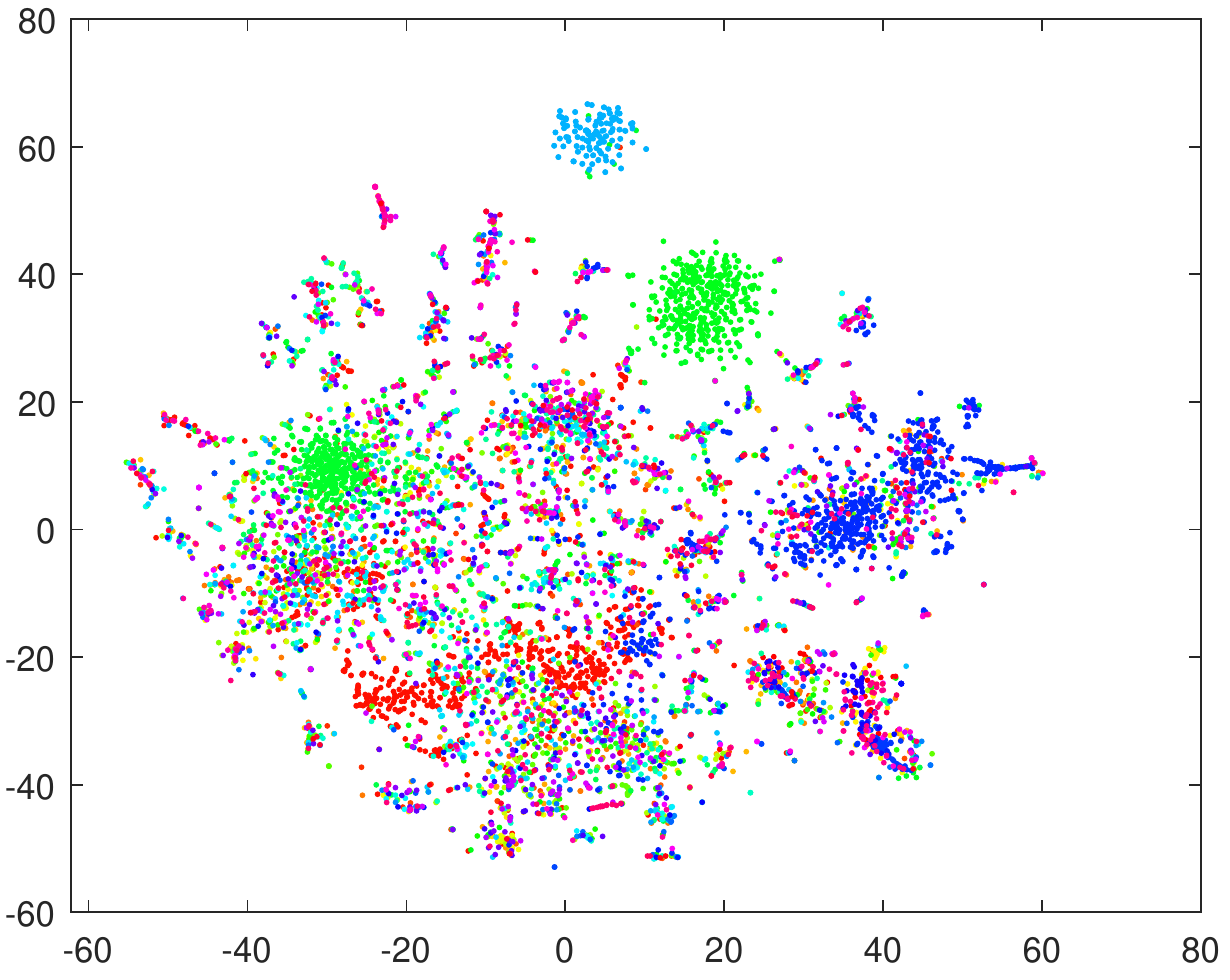}}}
	\caption{Visualization of the sample distributions from the $1$st view on the Caltech101 dataset by t-SNE. Features in the $1$st view are extracted by the MSD \cite{liu2011image}. (a) utilizes t-SNE and visualizes all original features directly. (b) and (c) are first preprocessed by KMSA-PCA and KMSA-LDA to obtain the 30-dimensional representations. Then, these representations are processed and visualized by t-SNE. It is clear that the distributions of most samples from the original data are disordered. KMSA-PCA and KMSA-LDA can achieve better performances. Especially after using KMSA-LDA, samples from the $1$st view are separated into many clusters, which is helpful for many applications. }
	\label{fig9}
\end{figure*}

It can be found in Table \ref{tab6} that the self-weighting scheme can help KMSA achieve a better performance than that obtained simply by fixing all the weights to $1/m$. It proves the effectiveness of the learned weights for KMSA. Additionally, the performances of the two kernel DR methods are not as good as that of KMSA-PCA, verifying that KMSA is a better method. 

\subsection{Visualization of KMSA}

To visualize the sample distribution of the low-dimensional features learned by KMSA, we adopt the t-distributed stochastic neighbor embedding (t-SNE) \cite{maaten2008visualizing} to embed all the learned features into a 2-dimensional subspace and visualize their distributions. This experiment is conducted on the Caltech101 dataset, and we visualize the learned features from the $1$st view. Beforehand, we utilize KMSA-PCA and KMSA-LDA to conduct multiview learning and project the original features from the $1$st view into a 30-dimensional subspace by considering the information from all views. The visualization of the feature distribution from the $1$st view on Caltech101 is shown in Fig. \ref{fig9}:

Even though the sample distributions of some specific classes from the original data are relatively concentrated, the distributions of most samples are disordered. It is clear that KMSA-PCA and KMSA-LDA can achieve better performances. Especially after KMSA-LDA is conducted, samples from the $1$st view are separated into many clusters, which is helpful for many applications.

\section{Conclusions}
In this paper, we proposed a generalized multiview graph embedding framework named kernelized multiview subspace analysis (KMSA). KMSA addresses multiview data in the kernel space to fully exploit the data representations within multiviews. It adopts a co-regularized term to minimize the divergence among views and utilizes a self-weighted strategy to learn the weights for all views, combining self-weighted learning with the co-regularized term, to deeply exploit the information from multiview data. We conducted various experiments on 6 datasets for multiview data classification and image retrieval. The experiments verified that KMSA is superior to the other multiview-based methods.

\section*{Acknowledgments}

We thank the anonymous reviewers for their valuable comments and suggestions, which enabled us to significantly improve the quality of this paper. This work is supported by the National Key Research and Development Program of China under grant 2018YFB0804203, by the National Natural Science Foundation of China (Grant 61806035, U1936217, 61672365, 61732008, 61725203, 61370142 and 62002041), by The Key Research and Technology Development Projects of Anhui Province (No. 202004a05020043), by the Postdoctoral Science Foundation (3620080307), by the Dalian Science and Technology Innovation Fund (2019J11CY001), and by the Liaoning Revitalization Talents Program (XLYC1908007).

\ifCLASSOPTIONcaptionsoff
  \newpage
\fi



%
\bibliographystyle{IEEEtran}
\bibliography{IEEEfull}

%

\begin{IEEEbiography}[{\includegraphics[width=1in,height=1.35in]{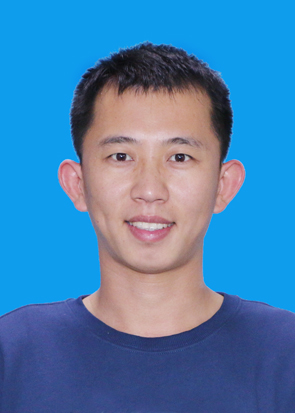}}]{Huibing Wang}
received a Ph.D. degree from the School of Computer Science and Technology, Dalian University of Technology, Dalian, in 2018. During 2016 and 2017, he was a visiting scholar at the University of Adelaide, Adelaide, Australia. Now, he is currently an Associate Professor at Dalian Maritime University, Dalian, Liaoning, China. He has authored and co-authored more than 40 papers in some famous journals or conferences. His research interests include computer vision and machine learning.
\end{IEEEbiography}

\begin{IEEEbiography}[{\includegraphics[width=1in,height=1.25in]{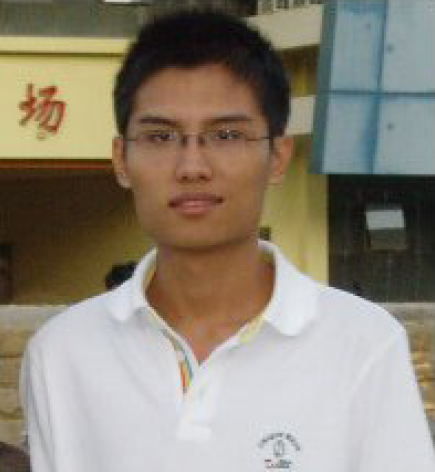}}]{Yang Wang}
earned a Ph.D. degree from The University of New South Wales, Kensington, Australia, in 2015. He is currently a Full Professor at Hefei University of Technology, China. He has published 50 research papers, most of which have appeared at competitive venues, including IEEE TIP, IEEE TNNLS, ACM TOIS, IEEE TCSVT, IEEE TMM, IEEE TCYB, IEEE TKDE, Pattern Recognition, Neural Networks, ACM Multimedia, ACM SIGIR, IJCAI, IEEE ICDM, ACM CIKM, and VLDB Journal. He currently serves as the Associate Editor of ACM Transactions on Information Systems (ACM TOIS), as a program committee member for IJCAI, AAAI, ACM Multimedia, ECMLPKDD, etc., and as the invited Journal Reviewer for more than 15 leading journals, such as IEEE TPAMI, IEEE TIP, IEEE TNNLS, ACM TKDD, IEEE TKDE, Machine Learning (Springer), and IEEE TMM. He received the best research paper runner-up award at PAKDD 2014.
\end{IEEEbiography}

\begin{IEEEbiography}[{\includegraphics[width=1in,height=1.25in]{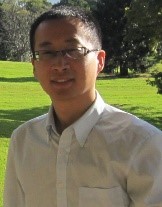}}]{Zhao Zhang}
 is a Full Professor in the School of Computer Science and School of Artificial Intelligence, Hefei University of Technology, Hefei, China. He received a Ph.D. degree from the Department of Electronic Engineering at City University of Hong Kong in 2013. Dr. Zhang was a Visiting Research Engineer at the National University of Singapore from February to May 2012. He then visited the National Laboratory of Pattern Recognition at the Chinese Academy of Sciences from September to December 2012. From October 2013 to October 2018, he was a Distinguished Associate Professor at the School of Computer Science, Soochow University, Suzhou, China. His current research interests include data mining and machine learning, image processing, and pattern recognition. He has authored/co-authored over 80 technical papers published at several prestigious journals and conferences, such as IEEE TIP (4), IEEE TKDE (5), IEEE TNNLS (4), IEEE TCSVT, IEEE TCYB, IEEE TSP, IEEE TBD, IEEE TII (2), ACM TIST, Pattern Recognition (6), Neural Networks (8), Computer Vision and Image Understanding, IJCAI, ACM MM, ICDM, ICASSP and ICMR. Dr. Zhang serves as an Associate Editor (AE) of Neurocomputing, IEEE Access and IET Image Processing. In addition, he has served as a Senior Program Committee (SPC) member or Area Chair (AC) of PAKDD, BMVC, and ICTAI and as a PC member for 10+ prestigious international conferences. He is now a Senior Member of the IEEE and a Member of the ACM.
\end{IEEEbiography}

\begin{IEEEbiography}[{\includegraphics[width=1in,height=1.35in]{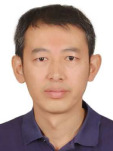}}]{Xianping Fu}
received a B.E. degree and an M.E. from Qiqihar Light Industry University in 1994 and 1997, respectively, and a Ph.D. degree from Dalian Maritime University in 2005. He was a postdoc at Tsinghua University in 2008 and at Harvard University in 2009. His major research interests are image processing for content recognition, multimedia technology, video processing for very low bit rate compression, a hierarchical storage management system for multimedia data, image processing for content retrieval, and driving and traffic environment perception.
\end{IEEEbiography}

\begin{IEEEbiography}[{\includegraphics[width=1in,height=1.25in]{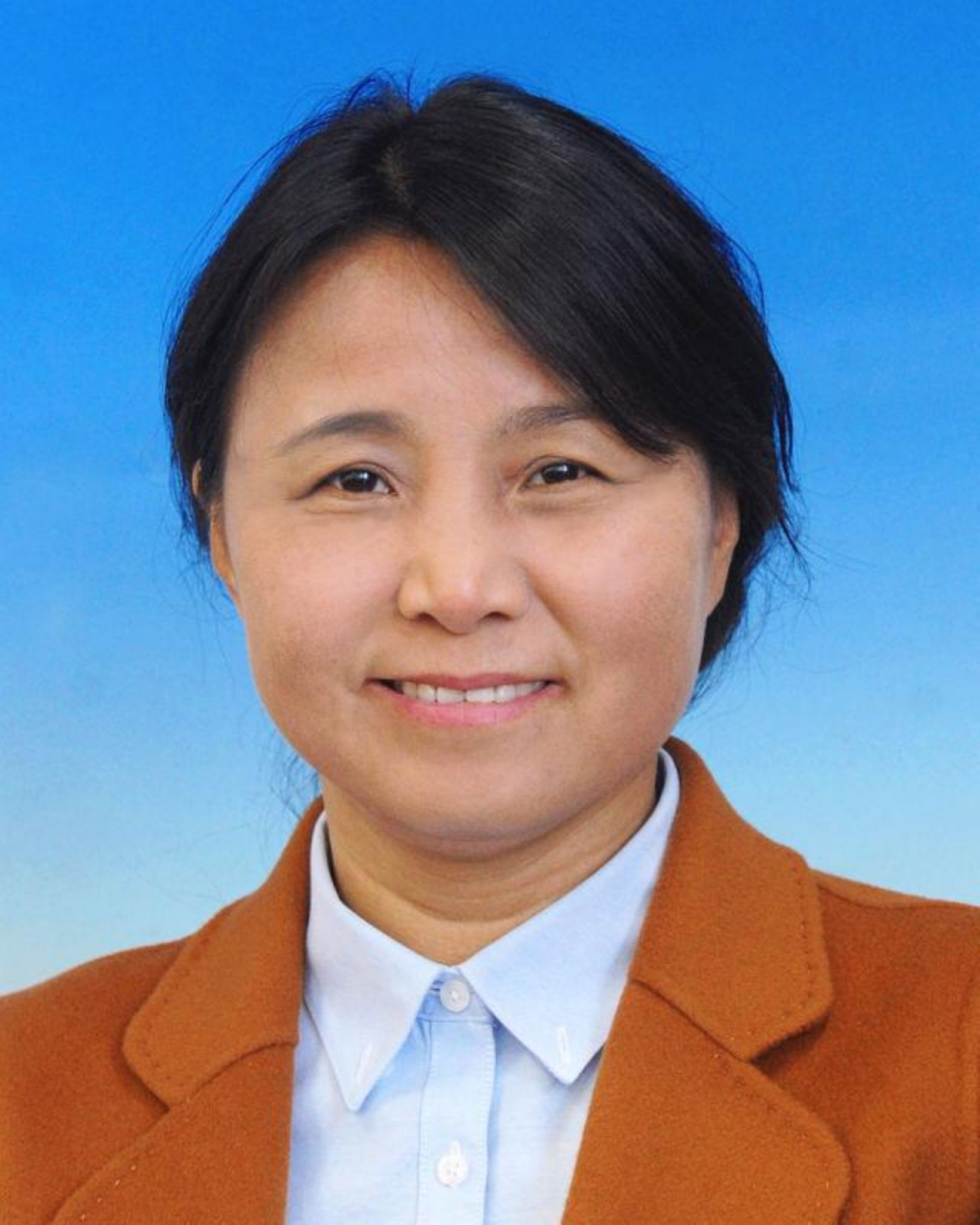}}]{Li Zhuo} received her B.E. degree in Radio Technology from the University of Electronic Science and Technology, Chengdu, China, in 1992, M.E. degree in Signal \& Information Processing from the Southeast University, Nanjing, in 1998, and Ph.D. degree in Pattern Recognition and Intellectual System from Beijing University of Technology, in 2004. She is currently a Professor, with Ph.D. supervision at Beijing University of Technology. During 2006-2007, she spent one-year as a Postdoctoral Research Fellow at the University of Sydney, Australia. Her current research interests include Multimedia Communication, image/video processing and Computer Vision as well. She has taken charge of over 40 research projects and has over 190 publications
\end{IEEEbiography}

\begin{IEEEbiography}[{\includegraphics[width=1in]{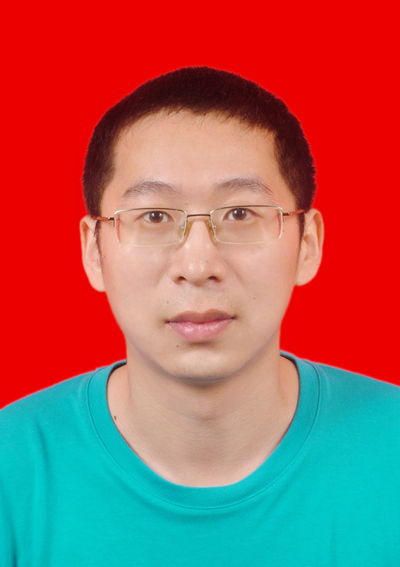}}]{Mingliang Xu} is a Full Professor at the School of Information Engineering of Zhengzhou University, China, and currently is the director of the CIISR (Center for Interdisciplinary Information Science Research) and the vice General Secretary of ACM SIGAI China. He received his Ph.D. degree in Computer Science and Technology from the State Key Laboratory of CAD$\&$CG at Zhejiang University, Hangzhou, China. His current research interests include computer graphics and artificial intelligence. He has authored more than 80 journal and conference papers in these areas, published in ACM TOG, ACM TIST, IEEE TPAMI, IEEE TIP, IEEE TCYB, IEEE TCSVT, IEEE TAC, IEEE TVCG, ACM SIGGRAPH (Asia), CVPR, ACM MM, IJCAI, etc.
\end{IEEEbiography}

\begin{IEEEbiography}[{\includegraphics[width=1in,height=1.25in]{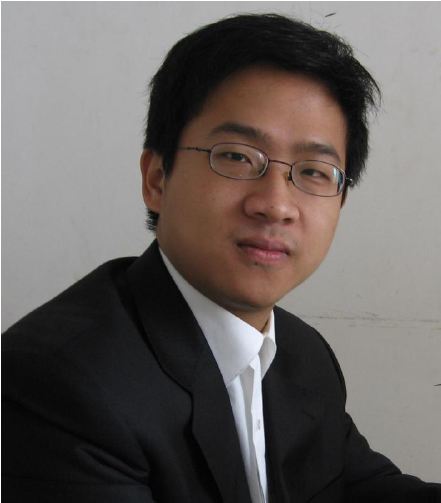}}]{Meng Wang}
is a Full Professor at Hefei University of Technology, China. He received a B.E. degree and Ph.D.
degree in the Special Class for the Gifted Young and in Signal
and Information Processing from the University of Science
and Technology of China, Hefei, China, respectively. He
previously worked as an associate researcher at Microsoft
Research Asia and then as a core member of a startup in the Bay
area. After that, he worked as a Senior Research Fellow at
the National University of Singapore. His current research interests include
multimedia content analysis, search, mining, recommendation, and large
-scale computing. He has authored 6 book chapters and over 100 journal and
conference papers in these areas, which have been published in TMM, TNNLS, TCSVT, TIP,
TOMCCAP, ACM MM, WWW, SIGIR, ICDM, etc. He has received paper
awards from ACM MM 2009 (Best Paper Award), ACM MM 2010 (Best Paper
Award), MMM 2010 (Best Paper Award), ICIMCS 2012 (Best Paper Award),
ACM MM 2012 (Best Demo Award), ICDM 2014 (Best Student Paper Award),
PCM 2015 (Best Paper Award), SIGIR 2015 (Best Paper Honorable Mention),
IEEE TMM 2015 (Best Paper Honorable Mention), and IEEE TMM 2016 (Best
Paper Honorable Mention). He is the recipient of the 2014 ACM SIGMM Rising Star
Award. He is/has been an Associate Editor of IEEE Trans. on Knowledge
and Data Engineering (TKDE),  IEEE Trans. Multimedia (TMM), IEEE Trans. on Neural Networks and Learning
Systems (TNNLS), and IEEE Trans. on Circuits and Systems for Video
Technology (TCSVT). He is a Senior Member of the IEEE.
\end{IEEEbiography}

\end{document}